 \documentclass[11pt,bezier,]{article}

\newtheorem{proposition}{Proposition}

\usepackage{amsmath}
\usepackage{amssymb}
\usepackage{amsfonts}
\usepackage{mathrsfs}
\usepackage{subfigure}
\usepackage{epsfig}
\usepackage{graphicx}
\usepackage{color}

\usepackage[linesnumbered,ruled,]{algorithm2e}

\newcommand{\myBox}{\squareforqed}

\newenvironment{proof}{\par\noindent{\bf Proof:}}{\hfill$\myBox$\par}

\newcommand{\blackhole}[1]{}

\newfont{\bbfv}{   msbm5}                      
\newfont{\bbfvi}{  msbm6}                      
\newfont{\bbfvii}{ msbm7}                      
\newfont{\bbfviii}{msbm8}                      
\newfont{\bbfix}{  msbm9}                      
\newfont{\bbfx}{   msbm10}                     
\newfont{\bbfxi}{  msbm10 scaled\magstephalf}  
\newfont{\bbfxii}{ msbm10 scaled\magstep1}     
\newfont{\bbfxiv}{ msbm10 scaled\magstep2}     
\newfont{\bbfxvii}{msbm10 scaled\magstep3}     
\newfont{\bbfxx}{  msbm10 scaled\magstep4}     
\newfont{\bbfxxv}{ msbm10 scaled\magstep5}     

\def\bbf#1{{\relax\rm
\ifdim\the\fontdimen6\the\font<7pt         
 \mbox{\bbfv #1}%
\else\ifdim\the\fontdimen6\the\font<7.6pt  
 \mbox{\bbfvi #1}%
\else\ifdim\the\fontdimen6\the\font<8.25pt 
 {\ifmmode\mathchoice{\mbox{\bbfvii #1}}
  {\mbox{\bbfvii #1}}{\mbox{\bbfvi #1}}
  {\mbox{\bbfv #1}}\else{\bbfvii #1}\fi}%
\else\ifdim\the\fontdimen6\the\font<8.85pt 
 {\ifmmode\mathchoice{\mbox{\bbfviii #1}}
  {\mbox{\bbfviii #1}}{\mbox{\bbfvi #1}}
  {\mbox{\bbfv #1}}\else{\bbfviii #1}\fi}%
\else\ifdim\the\fontdimen6\the\font<9.7pt  
 {\ifmmode\mathchoice{\mbox{\bbfix #1}}
  {\mbox{\bbfix #1}}{\mbox{\bbfvi #1}}
  {\mbox{\bbfv #1}}\else{\bbfix #1}\fi}%
\else\ifdim\the\fontdimen6\the\font<10.5pt 
 {\ifmmode\mathchoice{\mbox{\bbfx #1}}
  {\mbox{\bbfx #1}}{\mbox{\bbfvii #1}}
  {\mbox{\bbfv #1}}\else{\bbfx #1}\fi}%
\else\ifdim\the\fontdimen6\the\font<11.4pt 
 {\ifmmode\mathchoice{\mbox{\bbfxi #1}}
  {\mbox{\bbfxi #1}}{\mbox{\bbfviii #1}}
  {\mbox{\bbfvi #1}}\else{\bbfxi #1}\fi}%
\else\ifdim\the\fontdimen6\the\font<13pt   
 {\ifmmode\mathchoice{\mbox{\bbfxii #1}}
  {\mbox{\bbfxii #1}}{\mbox{\bbfviii #1}}
  {\mbox{\bbfvi #1}}\else{\bbfxii #1}\fi}%
\else\ifdim\the\fontdimen6\the\font<15pt   
 {\ifmmode\mathchoice{\mbox{\bbfxiv #1}}
  {\mbox{\bbfxiv #1}}{\mbox{\bbfx #1}}
  {\mbox{\bbfvii #1}}\else{\bbfxiv #1}\fi}%
\else\ifdim\the\fontdimen6\the\font<18pt   
 {\ifmmode\mathchoice{\mbox{\bbfxvii #1}}
  {\mbox{\bbfxvii #1}}{\mbox{\bbfxii #1}}
  {\mbox{\bbfx #1}}\else{\bbfxvii #1}\fi}%
\else\ifdim\the\fontdimen6\the\font<23pt   
 {\ifmmode\mathchoice{\mbox{\bbfxx #1}}
  {\mbox{\bbfxx #1}}{\mbox{\bbfxiv #1}}
  {\mbox{\bbfxii #1}}\else{\bbfxx #1}\fi}%
\else                                      
 {\ifmmode\mathchoice{\mbox{\bbfxxv #1}}
  {\mbox{\bbfxxv #1}}{\mbox{\bbfxx #1}}
  {\mbox{\bbfxvii #1}}\else{\bbfxxv #1}\fi}%
\fi\fi\fi\fi\fi\fi\fi\fi\fi\fi\fi}}

\begin{document}

\title{\Large\bf 
A Concave Optimization Algorithm for Matching Partially Overlapping Point Sets
} 

\date{}

\author{\begin{tabular}[t]{c@{\extracolsep{4em}}c@{\extracolsep{4em}}c} 
 Wei Lian &  Lei Zhang\\ \\
Dept. of Computer Science & Dept. of Computing \\
Changzhi University & The Hong Kong Polytechnic University  \\
Changzhi, Shanxi, China, 046031 & Hong Kong, China \\
E-mail: lianwei3@foxmail.com & 
\end{tabular}}

\maketitle


\thispagestyle{empty}
\subsection*{\centering Abstract}
{\em Point matching refers to  the process  of finding spatial transformation and correspondences between two sets of points.
In this paper,
we focus on the case that
there is only partial overlap between two point sets.
Following the approach of the robust point matching method,
we model point matching as a mixed linear assignment$-$least square problem
and show that after eliminating the transformation variable,
the resulting  problem of minimization with respect to point correspondence is a concave optimization problem.
Furthermore, this problem has the property that
the objective function can be converted into a form
with few nonlinear terms  via a linear transformation.
Based on these properties,
we employ the branch-and-bound (BnB) algorithm to  optimize the resulting problem
where the dimension of the search space is small.
To further improve  efficiency of the BnB algorithm 
where  computation of the lower bound is the bottleneck,
we propose a new lower bounding scheme
which has a k-cardinality linear assignment formulation 
and  can be efficiently solved. 
Experimental results show that 
the proposed  algorithm outperforms state-of-the-art methods in terms of robustness to disturbances 
and point matching accuracy.}

%
\section{Introduction}

Point matching refers to the process  of finding transformation and  correspondences between two sets of points.
It is a key component 
in many areas including 
computer vision, pattern recognition and medical image analysis
with applications such as structure-from-motion, tracking, 
image retrieval and object recognition \cite{book_multiple_view}. 
Disturbances such as deformation, positional noise, occlusion and outliers 
often makes  this problem  difficult. 

To address these difficulties,
many methods have been proposed 
(please refer to Sec. \ref{sect:related_work} for an overview).
Among them,
one of the most influential and successful is the robust point matching (RPM) method.
RPM models point matching as a mixed linear assignment$-$least square problem,
where the objective function is a cubic polynomial in its variables 
and is difficult to solve directly.
To address this difficulty,
RPM relaxes point correspondence to be fuzzily valued 
and employs the deterministic annealing (DA) technique \cite{deterministic_annealing} 
to gradually recover the  point correspondence.
But DA  is a heuristic scheme with no global optimality guarantee.
This means that RPM may performs poorly if encountering difficult matching problems.
Besides, 
DA initially matches points in two sets with equal chance, 
which causes the centers of mass of  two point sets to be aligned.
This trend may persist as algorithm iterates,
thus creating  a bias in favor of  matching the centers of  two point sets.

To address the problems associated with the use of DA in RPM,
in \cite{RPM_concave},
Lian and Zhang showed that under certain conditions,
the objective function of RPM can be converted into 
a concave quadratic function of point correspondence.
Besides, 
this function has 
a low rank Hessian matrix.
Based on these properties,
they then used the BnB algorithm for optimization
whose  search space has a dimension equal to the number of transformation parameters.
The resulting matching algorithm has several desirable advantages over  RPM. 
First, it is globally optimal and thus  more robust to disturbances such as extraneous structures; 
second, it can be rendered invariant to the corresponding transformation 
when simple transformations such as similarity  are employed. 
But the method assumes that each model point  has a counterpart in  scene  set.
This assumption no longer  holds in applications where there are outliers in both point sets.

To address this problem,
in \cite{RPM_model_occlude},
Lian and Zhang showed that
by relaxing the condition  that each model point  has a counterpart in  scene set,
the objective function of RPM can still be converted into a concave function of point correspondence,
which, albeit not being quadratic,
still has a low rank structure.
Therefore,
it is tractable to use the BnB algorithm  for optimization.
They also proposed a new lower bounding scheme
where the lower bounding problem has a  k-cardinality linear assignment formulation
and can be efficiently solved.

This paper is an extended version of  \cite{RPM_model_occlude}
where we make the following contributions:
First,
all the transformation parameters including those of translation are regularized in \cite{RPM_model_occlude}
which causes the method not to be  translation invariant,
whereas translation invariance is certainly a desirable property for a matching algorithm.
To address this problem,
we show that the objective function of RPM is a strictly convex quadratic function of translation
regardless of the values of other variables.
Therefore, translation can first be eliminated via minimization.
Then, 
we only need to enforce  regularization on the non-translational part of the transformation.
This enables  our new  matching algorithm to be translation invariant
and thus more applicable to practical  problems.

Second,
the method in \cite{RPM_model_occlude} uses regularization on transformation  
which forces the transformation solution to be close to a predefined value.
But in practice,
one may encounter the situation that 
the actual transformation deviates significantly from the predefined value
and thus  matching may fail as a result.
To address this problem,
we propose a new formulation targeted at the 2D/3D similarity registration problems,
where regularization on transformation is abandoned in favor of constraints on transformation.
This results in a matching algorithm  invariant to similarity transformation.
The new optimization problem is still concave and has a low rank structure.
Therefore, 
it is tractable to use the BnB algorithm for optimization.

The remainder  of the paper is organized as follows. 
We first review related work in Sec. \ref{sect:related_work} and RPM  in Sec. \ref{sect:RPM_objective}. 
We then discuss the new energy functions and their optimization 
in Sec. \ref{sect:energy} and \ref{sect:optimize}, respectively. 
We finally present the experimental results in Sec. \ref{sec:exp}
and conclude the paper in Sec. \ref{sect:conclude}.

\section{Related work\label{sect:related_work}}

\subsection{Heuristic based methods} 
The  category  of methods closely related with our method 
are those modeling both spatial transformation and point correspondence.
The  iterative closest point (ICP) method \cite{ICP,ICP2} 
is a well known  point matching method due to its simplicity and speed.
ICP  iterates between recovering  point correspondence based on  nearest neighbor relationship and
updating transformation as a least square problem.
But ICP is prone to be trapped in local minima 
because of  the discrete nature of point correspondence.
To address this problem,
the robust point matching (RPM) method \cite{RPM_TPS} 
relaxes point correspondence to be fuzzily   valued and
uses deterministic annealing (DA) to gradually recover point correspondence.
But  DA has the tendency of leading to  matching results
where the centers of mass of two point sets are aligned.
To address this problem,
the covariance matrix of the transformation parameters is used to
guide the determination of point correspondence in \cite{CDC},
which results in  better robustness to missing or extraneous structures.
But because the size of the covariance matrix is square times the number of transformation parameters,
this method is only suitable for transformations with few parameters.
Recently, the $L_2E$ estimator is introduced into point matching in \cite{L2E_mismatch}
to more robustly estimate the spatial transformation.
What's  common  with the above methods is that they are all heuristically based.
Therefore,  these methods may fail if the employed heuristics don't fit the matching problem.

The second category of methods are those modeling only spatial transformation.
Most of these methods are based on the idea that
a point set can be viewed  as the result of sampling from a distribution.
Among these methods,
the coherent point drift  method \cite{CPD} casts the point matching problem as that of
fitting a Gaussian Mixture Model (GMM)  representing one point set to another point set.
The expectation-maximization algorithm is used for optimization.
To eliminate the need for solving for point  correspondence,
Glaunes \textit{et al.} \cite{diffeo_match} formulate point matching as 
aligning two weighted sum of Dirac measures representing two point sets.
But  Dirac measures  is difficult to numerically compute.
To address this problem,
in \cite{kernel_Gaussian_journal},
two GMMs are used to represent two point sets  and the $L_2$ distance between them is minimized.
The early proposed kernel correlation method of \cite{kernel_correlation}
can be viewed as a special case of \cite{kernel_Gaussian_journal}.
The method of \cite{kernel_Gaussian_journal} was later improved in \cite{implicit_correspondence} 
by using the log-exponential function.
Under this formulation, ICP can be interpreted as a special case.
The method of \cite{kernel_Gaussian_journal} was also generalized to solve the group-wise point set registration problem
in \cite{kernel_groupwise, kernel_groupwise_IJCV}.
Recently, 
the Schr\"odinger distance transform  is used to represent point sets in \cite{SDT_match}
and  the point set registration problem is converted into  the problem of
 computing the geodesic distance between two points on a unit Hilbert sphere.
A common problem with the above methods is that since point correspondence is not modeled in these methods,
the one-to-one correspondence constraint is not enforced.
Therefore,
these methods tend to yield inferior matching results than those  enforcing the 
constraint
when encountering difficult matching problems such as the one as studied 
in this paper, i.e., where there is only partial overlap between two point sets.

The third category of methods are those  modeling only point correspondence.
Graph matching is used to solve the registration problem in \cite{GM_relax_label, GM_relax_label2}
by using the relaxation labeling technique.
But the methods need to be initialized by using features such as shape context \cite{SC} and SIFT \cite{SIFT}.


\subsection{Globally optimal methods}
Instead of solving the difficult problem of aligning two distributions representing two point sets,
Ho \textit{et al.} \cite{algebraic_moment}  proposed to match the moments of  distributions.
This   results in a system of polynomial equations
which can  be solved by algebraic geometric techniques.
But due to use of moments,
the method is  sensitive to occlusions and outliers.
Maciel and Costeira \cite{correspondence_concave} proposed a general framework 
to convert any correspondence problem into a concave optimization problem.
But the resulting concave problem is still hard 
and the optimization techniques employed there are only suitable for small scale problems.

The branch-and-bound (BnB) algorithm is a popular global optimization technique widely used in computer vision.
It is used in \cite{Lipschitz_3D_align} to align two sets of 3D shapes
based on the Lipschitz optimization theory.
But the method does not permit the presence of occlusion or outliers. 
BnB is used to recover 3D rigid transformation in \cite{branch_bound_align}.
But the 
correspondence needs to be known a priori,
which limits the applicability  of the method.
BnB is applied to optimize the RPM objective function in \cite{B_B_RPM},
where branching over the correspondence variable 
and over the transformation variable are both considered.
But due to lack of good  structures for optimization,
the proposed methods are only suitable for small scale problems. 
BnB is used to optimize the ICP objective function in \cite{Go-ICP}
by exploiting the special structure of the geometry of 3D rigid motions. 
BnB was recently  applied to  the problem of  
consensus set maximization (CSM) \cite{BnB_consensus_integer},
which seeks the best transformation maximizing the number of inliers.
The CSM framework was used for the correspondence and grouping problems 
in \cite{BnB_consensus_group}.
But 
the method is quite slow due to lack of efficient optimization techniques 
for the resulting bounding problems.
In the case that there is only 3D rotation between two point sets,
efficient methods have been proposed \cite{BnB_consensus_rotate,BnB_consensus_project}.
But the success of these methods critically depend on if the estimated translations are correct.
\section{
The energy function of RPM 
\label{sect:RPM_objective}}
Since our energy function originates from the energy function of RPM \cite{RPM_TPS}, 
we will first briefly review RPM. 
Suppose  we are two point sets in $\mathbb R^d$ to be matched: 
the model  set 
$\mathscr{X}=\{ \mathbf x_i,i=1,\ldots,m\}$ with  point 
$ \mathbf x_i=\begin{bmatrix}x_i^1,\ldots,x_i^d\end{bmatrix}^\top$,
and the scene set 
$\mathscr{Y}=\{ \mathbf y_j,j=1,\ldots,n\}$ with  point 
$ \mathbf y_j=\begin{bmatrix}y_j^1,\ldots,y_j^d \end{bmatrix}^\top$. 
To solve this problem, 
RPM jointly estimates  transformation and  point correspondence. 
It models point matching as a mixed linear assignment$-$least square problem:
\begin{subequations}
\begin{align} 
\min \quad & \widetilde E (\mathbf P,{\boldsymbol\vartheta})= 
\sum_{i,j}p_{ij}\| \mathbf y_j-T( \mathbf x_i| {\boldsymbol\vartheta})\|^2  
+ g({\boldsymbol\vartheta})  \label{energy0}  \\ 
s.t.\quad &
\mathbf P \mathbf 1_n\le  \mathbf 1_m, \quad \mathbf 1_m^\top \mathbf P\le \mathbf 1_n^\top,   
\quad  p_{ij}\in\{0,1\} \label{P_constraint0}
\end{align}       
\end{subequations}
where  $\mathbf P=\{p_{ij}\}$ is the correspondence matrix
with  $p_{ij}=1$ if there is a matching between $\mathbf x_i$ and $\mathbf y_j$ and $0$ otherwise.
$\mathbf 1_m$ denotes the $m$-dimensional  vector of  all ones.
$T(\cdot|\boldsymbol\vartheta)$ is the spatial transformation 
with  parameters ${\boldsymbol\vartheta}$.
$g({\boldsymbol\vartheta})$ is a regularizer  on ${\boldsymbol\vartheta}$. 
To solve problem \eqref{energy0},  \eqref{P_constraint0},
RPM relaxes  the binary constraint $p_{ij}\in\{0,1\}$ 
to $0\le p_{ij}\le 1$ 
and 
employs deterministic annealing (DA) for optimization. 
However, 
DA is a heuristic scheme which causes RPM to be less robust to disturbances. 
In the next section, 
we will present a new energy function based on the objective function of  RPM, 
which is more amenable to global optimization.

\section{
The new energy function 
\label{sect:energy}}

\addtocounter{MaxMatrixCols}{12}

To make our problem tractable,
we restrict the type of transformations
to be the one capable of being decomposed as 
a translational part $\mathbf t$
plus a non-translational part $\boldsymbol\phi({\mathbf x}_i|\boldsymbol\theta)$,
i.e.,  
$T({\mathbf x}_i|{\boldsymbol\theta},{\mathbf t})= \boldsymbol\phi({\mathbf x}_i|\boldsymbol\theta) + {\mathbf t} $,
where  $\boldsymbol\theta$ are parameters for the non-translational part of the transformation.

Following the approach of RPM,
we model  point matching as 
a mixed linear assignment$-$least square problem,
\begin{align}
\min\ \widetilde E (\mathbf P,{{\boldsymbol\theta}},{\mathbf t})= &
\sum_{i,j}p_{ij}\| {\mathbf y}_j- \boldsymbol\phi( {\mathbf x}_i | \boldsymbol\theta) - {\mathbf t}\|^2_2  
   \notag \\
=& {\mathbf 1}^\top \mathbf P\mathbf{\widetilde y}  
+ \boldsymbol\phi^\top(\boldsymbol\theta) [\text{diag}(\mathbf P{\mathbf 1})\otimes \mathbf I_d]
\boldsymbol\phi(\boldsymbol\theta)
  - 2 \boldsymbol\phi^\top(\boldsymbol\theta)  (\mathbf P\otimes \mathbf I_d) \mathbf y \notag\\
&  + n_p \|{\mathbf t}\|^2_2
 -2 {\mathbf t}^\top [({\mathbf 1}^\top \mathbf P)\otimes \mathbf I_d] \mathbf y
 +2{\mathbf t}^\top [({\mathbf 1}^\top {\mathbf P}^\top)\otimes \mathbf I_d] \boldsymbol\phi(\boldsymbol\theta)  \label{E_lin_tr} \\
 \text{s.t.} \ \mathbf P  {\mathbf 1}\le  {\mathbf 1}_m, 
 {\mathbf 1}^\top &\mathbf P\le  {\mathbf 1}_n^\top,
 {\mathbf 1}^\top \mathbf P  {\mathbf 1}=n_p,  
\mathbf P\ge 0 
\label{k_card_P_const}
\end{align}
Here the vectors $\boldsymbol\phi(\boldsymbol\theta)=\begin{bmatrix}
                       \boldsymbol\phi^\top(\mathbf x_1|\boldsymbol\theta),\ldots,\boldsymbol\phi^\top(\mathbf x_m|\boldsymbol\theta)
                      \end{bmatrix}^\top$,
$\mathbf y\triangleq\begin{bmatrix}
   \mathbf y_1^\top, \dots, \mathbf y_n^\top
  \end{bmatrix}^\top$
and $\mathbf {\widetilde y}\triangleq \begin{bmatrix}
   \|\mathbf y_1\|_2^2, \dots, \|\mathbf y_n\|_2^2
  \end{bmatrix}^\top$.
 $\text{diag}(\cdot)$ denotes converting a vector into a diagonal matrix,
$\mathbf I_d$ denotes the $d$-dimensional identity matrix and $\otimes$ denotes the Kronecker product.

Constraint \eqref{k_card_P_const} means that the matching
 is  one-to-one.
To make our problem tractable,
in this paper, 
we also require that the  number of matches is a priori known to be  $n_p$, a constant positive integer.
 Constraint  \eqref{k_card_P_const} satisfies 
the total unimodularity property \cite{correspondence_concave,book_comb_optimize}, 
which means that the vertices of the polytope 
(i.e., bounded polyhedron) determined by \eqref{k_card_P_const} have integer valued coordinates. 

From 
Eq. \eqref{E_lin_tr},
one can see   that 
$\widetilde E$ is  strictly convex quadratic with respect to ${\mathbf t}$
regardless of the  values of other variables. 
Therefore,
the optimal ${\mathbf t^*}$ minimizing $E$ 
can be obtained via solving the equation $\frac{\partial \widetilde E}{\partial {\mathbf t}}=0$.
The result is: 
\[
 {\mathbf t^*}
 =\frac{1}{n_p}\{[ ({\mathbf 1}^\top\mathbf P)\otimes \mathbf I_d ] \mathbf y 
 - [ ({\mathbf 1}^\top\mathbf P^\top)\otimes \mathbf I_d ] \boldsymbol\phi(\boldsymbol\theta) \}  \notag
\]
Substituting  $ {\mathbf t^*}$ back  into $\widetilde E$,
$\mathbf t$ is eliminated and 
we arrive at an energy function only in variables $\mathbf P$ and ${\boldsymbol\theta}$:
\begin{align}
 \widetilde E (\mathbf P,{\boldsymbol\theta})= 
\boldsymbol\phi^\top(\boldsymbol\theta) \widetilde A(\mathbf P) \boldsymbol\phi(\boldsymbol\theta) 
-2 \boldsymbol\phi^\top(\boldsymbol\theta)\widetilde{\mathbf b} (\mathbf P) 
 + {\mathbf 1}^\top\mathbf P \mathbf{\widetilde y}  
-\frac{1}{n_p} 
\|[({\mathbf 1}^\top\mathbf P) \otimes \mathbf I_d]\mathbf y \|^2 \label{eng_no_tran}
\end{align}
where
\begin{align}
\widetilde{A} (\mathbf P) &\triangleq   \text{diag}(\mathbf P\mathbf 1)\otimes \mathbf I_d    
 -\frac{1}{n_p}  [(\mathbf P\mathbf 1)\otimes \mathbf I_d][(\mathbf 1^\top \mathbf P^\top)\otimes \mathbf I_d],   \notag\\
\widetilde{\mathbf b} (\mathbf P)  &\triangleq   (\mathbf P\otimes \mathbf I_d)\mathbf y
-\frac{1}{n_p}  [(\mathbf P\mathbf 1)\otimes \mathbf I_d][(\mathbf 1^\top \mathbf P)\otimes \mathbf I_d]\mathbf y   \notag
\end{align}

We next aims to eliminate $\boldsymbol\theta$ so as to
obtain   an energy function only in one variable $\mathbf P$.
We consider two ways  to achieve  such a goal   in this paper:
1) adding a regularizer on $\boldsymbol\theta$ 
so as to make  the energy function   a convex function of  $\boldsymbol\theta$.
Then,
$\boldsymbol\theta$ can be eliminated via convex optimization.
2) 
using   constraints on $\boldsymbol\theta$.
We will detail these two approaches in the following subsections.

\subsection{Approach  one: using  regularization on $\theta$  \label{subsec:scheme1}}

To make our problem tractable,
we restrict the type of transformations to be the one
whose non-translational part is linear with respect to its parameters,
i.e.,  $\boldsymbol\phi(\mathbf x_i|\boldsymbol\theta)=\mathbf J(\mathbf x_i) \boldsymbol\theta$. 
We consider the following form of regularization on $\boldsymbol\theta$:
$(\boldsymbol\theta-\boldsymbol\theta_0)^\top\mathbf H(\boldsymbol\theta-\boldsymbol\theta_0) - \boldsymbol\theta_0\mathbf H\boldsymbol\theta_0 $,
i.e., 
${\boldsymbol\theta}$ is required to be close to a predefined constant value  ${\boldsymbol\theta}_0$. 
Here $\mathbf H$ is a predefined constant symmetric weighting matrix. 
Under these requirements,
the energy function \eqref{eng_no_tran} becomes:

\begin{align}
 E(\mathbf P,{\boldsymbol\theta}) &
=\widetilde E (\mathbf P,{\boldsymbol\theta})|_{\boldsymbol\phi(\mathbf x_i|\boldsymbol\theta)=\mathbf J(\mathbf x_i) \boldsymbol\theta}
+ \boldsymbol\theta^\top \mathbf H\boldsymbol\theta -2 \boldsymbol\theta_0^\top\mathbf H\boldsymbol\theta \notag\\
= &{\boldsymbol\theta}^\top [A(\mathbf P)+\mathbf H] {\boldsymbol\theta} 
-2 {\boldsymbol\theta}^\top [\mathbf b(\mathbf P) + \mathbf H\boldsymbol\theta_0]
 + {\mathbf 1}^\top\mathbf P \mathbf{\widetilde y} 
 -\frac{1}{n_p} 
 \|[({\mathbf 1}^\top\mathbf P) \otimes \mathbf I_d]\mathbf y \|^2 \label{eng_reg}
\end{align}
where
\begin{align*}
A(\mathbf P)\triangleq & 
 \mathbf J^\top \{ \text{diag}(\mathbf P\mathbf 1)\otimes \mathbf I_d 
 -\frac{1}{n_p}  [(\mathbf P\mathbf 1)\otimes \mathbf I_d][(\mathbf 1^\top \mathbf P^\top)\otimes \mathbf I_d] \}\mathbf J,   \\
\mathbf b(\mathbf P)  \triangleq & \mathbf J^\top (\mathbf P\otimes \mathbf I_d) \mathbf y
-\frac{1}{n_p} \mathbf J^\top [(\mathbf P\mathbf 1)\otimes \mathbf I_d][(\mathbf 1^\top \mathbf P)\otimes \mathbf I_d] \mathbf y   
\end{align*}
Here the matrix  
$\mathbf J\triangleq\begin{bmatrix}
   \mathbf J^\top(\mathbf x_1), \dots, \mathbf J^\top(\mathbf x_m)
  \end{bmatrix}^\top$.

\begin{figure}[h]
\centering
\subfigure{\includegraphics[width=0.4\linewidth]{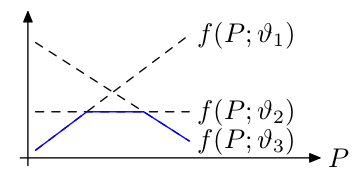}}

\caption{ 
Pointwise minimization of a family of linear functions (dashed straight  lines) 
results in a concave function (piecewise linear solid blue  line).
\label{pointwise_min}
}
\end{figure}

Under the condition that $A(\mathbf P)+\mathbf H$ is positive definite,
(denoted by $A(\mathbf P)+\mathbf H\succ 0$),
$E$ becomes a strictly convex quadratic function of $\boldsymbol\theta$
and the optimal $\boldsymbol\theta^*$ minimizing $E$ 
can be obtained via solving  the equation $\frac{\partial E}{\partial \boldsymbol\theta}=0$.
The result is
\[
 \boldsymbol \theta^*=(A(\mathbf P)+\mathbf H)^{-1} (b(\mathbf P)+\mathbf H\boldsymbol\theta_0)
\]
Substituting $\boldsymbol \theta^*$ back into Eq. \eqref{eng_reg},
$\boldsymbol\theta$ is eliminated and we arrive at an energy function only in one variable $\mathbf P$,
\begin{align}
 E(\mathbf P)
= & -(b(\mathbf P)^\top + \boldsymbol\theta_0^\top\mathbf H)(A(\mathbf P)+\mathbf H)^{-1}  (b(\mathbf P) + \mathbf H\boldsymbol\theta_0)  \notag\\
 & -\frac{1}{n_p}
\| [({\mathbf 1}^\top\mathbf P) \otimes \mathbf I_d]\mathbf y \|^2
+ {\mathbf 1}^\top\mathbf P \mathbf{\widetilde y} 
\end{align}
$E$ can be characterized by the following propositions:
\begin{proposition} \label{prop:conv_reg} 
 $E(\mathbf P)$ is concave over the spectrahedra
 $A(\mathbf P)+\mathbf H\succ0$.
 \end{proposition}
{\proof
Based on the preceding derivation,
we can see that 
$\widetilde E(\mathbf P,\boldsymbol\theta)=
\min_{\mathbf t} \widetilde E(\mathbf P,\boldsymbol\theta,\mathbf t)$ 
and $E(\mathbf P)=\min_{\boldsymbol\theta} \widetilde E(\mathbf P,\boldsymbol\theta)|_{\boldsymbol\phi(\mathbf x_i|\boldsymbol\theta)=\mathbf J(\mathbf x_i) \boldsymbol\theta}+ \boldsymbol\theta^\top \mathbf H\boldsymbol\theta -2\boldsymbol\theta_0\mathbf H\boldsymbol\theta$ 
when $A(\mathbf P)+\mathbf H\succ 0$.
Therefore we have 
$E(\mathbf P)=\min_{\mathbf t,\boldsymbol\theta} 
\widetilde E(\mathbf P,\boldsymbol\theta,\mathbf t)|_{\boldsymbol\phi(\mathbf x_i|\boldsymbol\theta)
=\mathbf J(\mathbf x_i) \boldsymbol\theta}+ \boldsymbol\theta^\top \mathbf H\boldsymbol\theta -2\boldsymbol\theta_0\mathbf H\boldsymbol\theta$
when $A(\mathbf P)+\mathbf H\succ 0$.
It is clear that 
$ \widetilde E(\mathbf P,\boldsymbol\theta,\mathbf t)|_{\boldsymbol\phi(\mathbf x_i|\boldsymbol\theta)=\mathbf J(\mathbf x_i) \boldsymbol\theta}$ is  linear with respect to  $\mathbf P$.
We see that $E(\mathbf P)$ is the result of point-wise minimization of a family of linear functions,
and hence is concave, as illustrated in Fig. \ref{pointwise_min}.

Since $n_p>0$, based on the property of  Schur complement, 
we have $A(\mathbf P)+\mathbf H\succ 0 \Leftrightarrow 
\begin{bmatrix}
   \mathbf J^\top [\text{diag}(\mathbf P\mathbf 1)\otimes \mathbf I_d ] J +\mathbf H & \mathbf J^\top [(\mathbf P\mathbf 1)\otimes \mathbf I_d] \\
   [(\mathbf 1^\top \mathbf P^\top)\otimes \mathbf I_d]J & n_p \mathbf I_d
  \end{bmatrix}\succ 0
$ where the latter inequality is a spectrahedra.

}

\begin{proposition} \label{prop_2}
There exists a minimum binary solution of
$E(\mathbf P)$  under  constraint \eqref{k_card_P_const} when $A(\mathbf P)+\mathbf H\succ0$.
\end{proposition}
{\proof
We already proved that $E$ is concave when $A(\mathbf P)+\mathbf H\succ0$.
 It is well known that the minimum solution of a concave
function over a polytope can be taken at one of its vertices.
The proposition follows by combining this result with the
total unimodularity of  constraint \eqref{k_card_P_const} as stated previously.
}

To facilitate  optimization of $E$,
matrix $\mathbf P$ needs first to be  vectorized. 
Let us define the vectorization of a matrix as the concatenation of its rows
\footnote{This  is different from the conventional definition. }, 
denoted by $\text{vec}(\cdot)$. 
Let $\mathbf p\triangleq \text{vec}(\mathbf P)$. 
To obtain a new form of $E$ which has fewer nonlinear terms, 
we need some new denotations. 
Let
\begin{align*}
\text{vec}\{  \mathbf J^\top [\text{diag}(\mathbf P  \mathbf 1_n) \otimes \mathbf I_d]\mathbf J \} =
 \text{vec}\{ \mathbf J_2^\top [ (\mathbf P  \mathbf 1_n) \otimes \mathbf I_{n_\theta}] \} &=\mathbf B\mathbf p,\\
  \mathbf J^\top (\mathbf P\otimes \mathbf I_d)\mathbf y &=\mathbf C\mathbf p,\\
  \text{vec} \{ \mathbf J^\top[(\mathbf P\mathbf 1)\otimes \mathbf I_d] \} &=\mathbf D\mathbf p, \\
   [(\mathbf 1^\top\mathbf P)\otimes \mathbf I_d]\mathbf y &=\mathbf F\mathbf p
 \end{align*}
where $n_\theta$ denotes the dimension of ${\boldsymbol\theta}$ and 
$
 \mathbf J_2\triangleq\begin{bmatrix}
      \mathbf J(\mathbf x_1)^\top \mathbf J(\mathbf x_1),       \ldots,      \mathbf J(\mathbf x_m)^\top \mathbf J(\mathbf x_m)
     \end{bmatrix}^\top
$.
Based on the fact 
$\text{vec}(\mathbf M_1\mathbf M_2\mathbf M_3)
= (\mathbf M_1\otimes \mathbf M_3^\top)\text{vec}(\mathbf M_2)$
for any matrices $\mathbf M_1$, $\mathbf M_2$ and $\mathbf M_3$,
we have
\begin{align*}
\mathbf B&=(\mathbf J_2^\top\otimes \mathbf I_{n_\theta}) \boldsymbol\Psi^{m,1}_{n_\theta} (\mathbf I_m \otimes\mathbf 1_n^\top),\\
\mathbf C&=(\mathbf J^\top\otimes\mathbf y^\top)\boldsymbol\Psi^{m,n}_d, \\
\mathbf D&=(\mathbf J^\top\otimes \mathbf I_d) \boldsymbol\Psi^{m,1}_d (\mathbf I_m\otimes \mathbf 1_n^\top),\\
\mathbf F&=(\mathbf I_d\otimes \mathbf y^\top) \boldsymbol\Psi^{1,n}_d (\mathbf 1_m^\top\otimes \mathbf I_n)
\end{align*}
Here  the $mnd^2\times mn$ matrix
$
\boldsymbol\Psi^{m,n}_d\triangleq
   \mathbf I_m \otimes  \begin{bmatrix}
\mathbf I_n \otimes (\mathbf e_d^1)^\top,
\dots,
\mathbf I_n \otimes (\mathbf e_d^d)^\top
\end{bmatrix}^\top
$
satisfies $\text{vec}(\mathbf M_{m,n}\otimes \mathbf I_d)=\boldsymbol\Psi ^{m,n}_d \text{vec}(\mathbf M_{m,n})$
for any $m\times n$ matrix $\mathbf M_{m,n}$,
where  $\mathbf e_d^i$ denotes the  $d$-dimensional column vector 
with  the $i$-th entry being  $1$ and all other entries being   $0$.
$\boldsymbol\Psi^{m,n}_d$ is a large but sparse matrix and can be implemented  using  function $speye$ in Matlab.

With the above preparation, 
$E$  can be written in terms of vector $\mathbf p$ as:
\begin{align}
 E(\mathbf p)
= & -[b^\top(\mathbf p) + \boldsymbol\theta_0^\top\mathbf H][\mathbf A(\mathbf p)+\mathbf H]^{-1} [b(\mathbf p)
+ \mathbf H\boldsymbol\theta_0] 
-\frac{1}{n_p}\| \mathbf F\mathbf p \|^2
 + ({\mathbf 1}^\top \otimes  \mathbf{\widetilde y}^\top) \mathbf p
   \label{eng_reg_vec}
\end{align}
where
\begin{align*}
\mathbf A(\mathbf p)\triangleq &
\text{mat}(\mathbf B\mathbf p)-
\frac{1}{n_p} \text{mat}(\mathbf D\mathbf p) \text{mat}^\top(\mathbf D\mathbf p) \\
\mathbf b(\mathbf p)  \triangleq &
\mathbf C\mathbf p-\frac{1}{n_p} \text{mat}(\mathbf D\mathbf p) \mathbf F\mathbf p
\end{align*}
Here $\text{mat}(\cdot)$ denotes reconstructing a symmetric matrix 
from a vector which is the  result of applying $\text{vec}(\cdot)$ to a symmetric matrix.
Therefore, $\text{mat}(\cdot)$ can be seen as 
the inverse of operator $\text{vec}(\cdot)$ applied to a symmetric matrix
and its meaning will be clear from the context.

Since $\mathbf  1^\top_{mn} \mathbf  p=n_p$, a constant value,
regardless of the values of $\mathbf p$,
rows in $\mathbf B$, $\mathbf C$, $\mathbf D$ and $\mathbf F$
equal to multiple of $\mathbf  1^\top_{mn}$ will be useless
and can be removed.
It can be verified that for
2D similarity and affine transformations and 3D scaling + translation transformation
(please refer to Sec. \ref{sec:exp} for detail),
$\mathbf B$ and $\mathbf D$ contains such rows.
Also, redundant rows can  be removed.
Since $\text{mat}(\mathbf B\mathbf p)$ and $\text{mat}(\mathbf D\mathbf p)$ are symmetric matrices,
$\mathbf B$ and $\mathbf D$ will  contain redundant rows.
Based on the above analysis,
We hereby  denote $\mathbf B_2$ (respectively $\mathbf D_2$) 
as the matrix formed as a result of $\mathbf B$ (respectively $\mathbf D$) removing such  rows.
Let the QR factorization of $
\begin{bmatrix}
\mathbf  B_2^\top, \mathbf  D_2^\top, \mathbf C^\top, \mathbf F^\top
\end{bmatrix}
$ be
$
\mathbf Q\mathbf \Gamma=\begin{bmatrix}
 \mathbf B_2^\top , \mathbf D_2^\top, \mathbf C^\top, \mathbf F^\top
\end{bmatrix}
$,
where $\mathbf \Gamma$ is an upper triangular matrix 
and the columns of $\mathbf Q$ are orthogonal unity vectors.
In view of the form of $E$ in  \eqref{eng_reg_vec},
we can see  that the nonlinear part $E_{c}$ of $E$ 
(i.e., all the terms except for the last  term in \eqref{eng_reg_vec})  
is   determined by variable
$\begin{bmatrix}
 \mathbf B_2^\top , \mathbf D_2^\top,  \mathbf C^\top, \mathbf F^\top
\end{bmatrix}^\top\mathbf p = \mathbf \Gamma^\top \mathbf Q^\top\mathbf p = \mathbf \Gamma^\top\mathbf u$,
which in turn is determined by a low dimensional variable $\mathbf u\triangleq  \mathbf Q^\top\mathbf p$.

The specific form of $E_{c}$ in terms of variable $\mathbf u$ is:
\begin{align}
 E_{c}(\mathbf u)
= &-[\mathbf b^\top(\mathbf u) + \boldsymbol\theta_0^\top\mathbf H]
[\mathbf A(\mathbf u)+\mathbf H]^{-1} [\mathbf b(\mathbf u) + \mathbf H\boldsymbol\theta_0] 
-\frac{1}{n_p}\| (\mathbf \Gamma^\top\mathbf u)_F \|^2
   \label{eng_reg_u}
\end{align}
where
\begin{align*}
\mathbf A(\mathbf u)\triangleq &
\text{mat}[(\mathbf \Gamma^\top\mathbf u)_{B_2}]
-\frac{1}{n_p} \text{mat}[(\mathbf\Gamma^\top\mathbf u)_{D_2}] \text{mat}^\top[(\mathbf\Gamma^\top\mathbf u)_{D_2}] \\
\mathbf b(\mathbf u)  \triangleq &
(\mathbf \Gamma^\top\mathbf u)_C -\frac{1}{n_p} \text{mat}[(\mathbf \Gamma^\top\mathbf u)_{D_2}] (\mathbf \Gamma^\top\mathbf u)_F
\end{align*}
Here $(\mathbf \Gamma^\top\mathbf u)_{B_2}$ 
denotes the vector formed by the elements of vector $\mathbf \Gamma^\top\mathbf u$
with  indices  equal to row indices of the submatrix $\mathbf B_2$ 
in matrix $\begin{bmatrix}
 \mathbf B_2^\top,\mathbf D_2^\top,\mathbf C^\top,\mathbf F^\top
  \end{bmatrix}^\top$.
Vectors $(\mathbf \Gamma^\top\mathbf u)_{D_2}$,
$(\mathbf \Gamma^\top\mathbf u)_{C}$ and $(\mathbf \Gamma^\top\mathbf u)_{F}$ are similarly defined.                                                                   
Here we abuse the use of 'mat' so that 
$\text{mat}(\mathbf B_2\mathbf p) =\text{mat}(\mathbf B\mathbf p)$
and $\text{mat}(\mathbf D_2\mathbf p) =\text{mat}(\mathbf D\mathbf p)$.
The meaning will be clear from the context.
Assume the numbers of rows in $\mathbf B_2$ and $\mathbf D_2$ are $n_{B_2}$ and $n_{D_2}$, respectively.
Then the dimension of $\mathbf u$ is $n_{B_2}+n_{D_2} + n_\theta +d$,
which is much smaller than that of $\mathbf p$ and also independent of the cardinalities  of the two point sets.
This is the key reason why  our algorithm is applicable to large scale problems
and  scale well with  problem size.

\subsection{Approach two: using constraints on $\theta$}
The advantage of the preceding approach is that
with $\boldsymbol\theta$ eliminated,
the subsequent optimization only involves $\mathbf P$
which results in good computational efficiency.
The disadvantage  is that
by using regularization on $\boldsymbol\theta$ where
prior information about $\boldsymbol\theta$ needs to be supplied,
the transformation solution is biased in favor of  the predefined value.
In particular,
the resulting point matching method is not rotation invariant. 
To address this problem,
in this section, instead of using regularization,
we will consider using  constraints  on $\boldsymbol\theta$.
However, with the increasing number of constraints,
the optimization problem becomes slower to solve.
Therefore, 
we will restrict the type of transformations to be the similarity transformation
whose number of constraints is small compared with other types of transformations.

To facilitate derivation of functions in the following,
we need to  rewrite   $\widetilde E$ in Eq. \eqref{eng_no_tran} 
using mainly  matrices instead of vectors.
It is easy to verify that   $\widetilde E$ can be 
rewritten as:
\begin{align}
 \widetilde E(\mathbf P,{\boldsymbol\theta})= &
\text{tr} \{\mathbf\Phi^\top(\boldsymbol\theta) [\text{diag}(\mathbf P\mathbf 1)
-\frac{1}{n_p} \mathbf P\mathbf 1 \mathbf 1^\top\mathbf P^\top]\mathbf\Phi(\boldsymbol\theta)\} \notag\\
& -2\text{tr}[\mathbf\Phi^\top(\boldsymbol\theta)
(\mathbf P-\frac{1}{n_p}\mathbf P\mathbf 1\mathbf 1^\top\mathbf P)\mathbf Y]
+\mathbf 1^\top\mathbf P\mathbf{\widetilde y} -\frac{1}{s}\|\mathbf 1^\top\mathbf P\mathbf Y\|^2  \label{eng_no_tran_mat}
\end{align}
where the matrices
$\mathbf\Phi(\boldsymbol\theta)=\begin{bmatrix}
                       \boldsymbol\phi(\mathbf x_1|\boldsymbol\theta),\ldots,\boldsymbol\phi(\mathbf x_m|\boldsymbol\theta)
                      \end{bmatrix}^\top$ and
$\mathbf Y=\begin{bmatrix}
    \mathbf y_1,\ldots,\mathbf y_n
   \end{bmatrix}^\top
$.
$\text{tr}(\cdot)$ denotes the trace of a matrix.

With the  transformation chosen as similarity:
 $\boldsymbol\phi(\mathbf x_i|\boldsymbol\theta)=s\mathbf R \mathbf x_i$,
where   $s$ denotes scale and $\mathbf R$ denotes   rotation matrix,
we have
$\mathbf\Phi(\boldsymbol\theta)=s \mathbf X \mathbf R^\top$,
where the matrix $\mathbf X\triangleq \begin{bmatrix}
    \mathbf x_1,\ldots,\mathbf x_m
   \end{bmatrix}^\top$.
Substituting this specification  into Eq. \eqref{eng_no_tran_mat},
we get our energy function as:
\begin{align}
& E(\mathbf P,s,\mathbf R)= \widetilde E(\mathbf P,\boldsymbol\theta)|_{\boldsymbol\phi(\mathbf x_i|\boldsymbol\theta)=s\mathbf R\mathbf x_i}  \notag\\
& = s^2 (\mathbf {\widetilde x}^\top\mathbf P\mathbf  1   - \frac{1}{n_p}    \| \mathbf X^\top\mathbf P\mathbf  1\|^2)  
 +\mathbf 1^\top\mathbf P \mathbf {\widetilde y} 
- 2s\,\text{tr} [ \mathbf R  \mathbf X^\top(\mathbf P - \frac{1}{n_p}\mathbf P\mathbf 1 \mathbf 1^\top\mathbf P) \mathbf Y]
 - \frac{1}{n_p} \|\mathbf  1^\top\mathbf P \mathbf Y\|^2
\end{align}
where the vector $\mathbf {\widetilde x}\triangleq \begin{bmatrix}
   \|\mathbf x_1\|_2^2, \dots, \|\mathbf x_{m}\|_2^2
  \end{bmatrix}^\top$.

It is clear that
\[
 \min_{\mathbf P,s,\mathbf R} E(\mathbf P,s,\mathbf R)=
 \min_{\mathbf P} \{\min_{s,\mathbf R} E(\mathbf P,s,\mathbf R)\} =\min_{\mathbf P} E(\mathbf P)
\]
where the energy function
\begin{gather}
 E(\mathbf P)\triangleq\min_{s,\mathbf R}  E(\mathbf P,s,\mathbf R) 
\label{E_P}
\end{gather}
Therefore,
the minimization of $E(\mathbf P,s,\mathbf R)$ now
boils down to the minimization of $E(\mathbf P)$.
$E(\mathbf P)$ can be characterized by the  following propositions:
\begin{proposition} \label{prop_3}
$E(\mathbf P)$ is concave. 
\end{proposition}
{\proof
Based on the aforementioned derivation of the energy function,
we have $\widetilde E(\mathbf P,\boldsymbol\theta )=\min_{{\mathbf t}} \widetilde E(\mathbf P,\boldsymbol\theta,\mathbf t)$ and
$E(\mathbf P) =\min_{s,\mathbf R} \widetilde E(\mathbf P,\boldsymbol\theta)|_{\boldsymbol\phi(\mathbf x_i|\boldsymbol\theta)=s\mathbf R\mathbf x_i}$.
Therefore, we have  \\
$E(\mathbf P)=\min_{s,\mathbf R,{\mathbf t}} \widetilde E(\mathbf P,\boldsymbol\theta,{\mathbf t})|_{\boldsymbol\phi(\mathbf x_i|\boldsymbol\theta)=s\mathbf R\mathbf x_i}$.
It is clear that
$\widetilde E(\mathbf P,\boldsymbol\theta,{\mathbf t})|_{\boldsymbol\phi(\mathbf x_i|\boldsymbol\theta)=s\mathbf R\mathbf x_i}$ is  linear with respect to $\mathbf P$.
We see that $E(\mathbf P)$ is the result of pointwise minimization of a family of linear functions,
and hence is concave, as illustrated in Fig. \ref{pointwise_min}.
}

\begin{proposition} \label{prop_4}
There exists a minimum binary solution of
$E(\mathbf P)$  under  constraint \eqref{k_card_P_const}.
\end{proposition}

We omit the proof of proposition \ref{prop_4} 
as it is similar to that of proposition \ref{prop_2}.
Proposition \ref{prop_3} shows that $E$ is concave regardless of the values of $\mathbf P$,
which is in contrast to the preceding approach
where  the energy function is only concave over a finite  region.

To facilitate  optimization of $E$,
$E$ needs to be expressed in terms of vector $\mathbf p$.
To obtain a new form of $E$ which has fewer nonlinear terms, 
we first need some  denotations. 
Let
\begin{align}
 \text{vec}(\mathbf X^\top\mathbf P\mathbf Y)\triangleq &\mathbf  B\mathbf  p, \label{B_denote}\\
 \mathbf X^\top\mathbf P\mathbf 1\triangleq & \mathbf C\mathbf  p, \\
 \mathbf Y^\top\mathbf P^\top\mathbf  1\triangleq & \mathbf D\mathbf p,\\
 \mathbf {\widetilde x}^\top\mathbf P \mathbf 1\triangleq & \mathbf  a^\top\mathbf  p
\end{align}
Based on the fact 
$\text{vec}(\mathbf M_1 \mathbf M_2 \mathbf M_3)
= (\mathbf M_1\otimes \mathbf M_3^\top)\text{vec}(\mathbf M_2)$
for any matrices $\mathbf M_1$, $\mathbf M_2$ and $\mathbf M_3$, we have
\[
\mathbf  B=\mathbf X^\top\otimes \mathbf Y^\top,
\mathbf C=\mathbf X^\top\otimes\mathbf  1_n^\top, 
\mathbf D=\mathbf 1_m^\top\otimes \mathbf Y^\top, 
\mathbf  a=\mathbf {\widetilde x} \otimes\mathbf  1_n
\]

With the above preparation, 
$E(\mathbf P)$  can be rewritten in terms of  $\mathbf p$ as
\begin{align}
E(\mathbf p)= &  
(\mathbf 1_m^\top\otimes\mathbf { \widetilde y}^\top)\mathbf  p 
- \frac{1}{n_p} \|\mathbf D\mathbf  p \|^2  
+ \min_{s,\mathbf R} \{s^2 (\mathbf a^\top\mathbf  p   - \frac{1}{n_p}    \| \mathbf C\mathbf  p \|^2)  \notag\\
 &- 2s\, \text{tr}(\mathbf R [ \text{mat} (\mathbf B\mathbf p)  - \frac{1}{n_p} \mathbf C\mathbf p\mathbf  p^\top \mathbf D^\top ]) \} \label{Eompact}
\end{align}

Let the QR factorization of matrix $
\begin{bmatrix}
\mathbf B^\top, \mathbf C^\top, \mathbf D^\top,\mathbf  a^\top
\end{bmatrix}
$ be
$
\mathbf  Q\mathbf  \Gamma=\begin{bmatrix}
\mathbf B^\top ,\mathbf C^\top,\mathbf D^\top,\mathbf  a^\top
\end{bmatrix}
$,
where $\mathbf \Gamma$ is an upper triangular matrix and 
the columns of $\mathbf Q$ are  orthogonal unity vectors.
In view of the form of $E$ in  \eqref{Eompact},
we can see  that the nonlinear part $E_{c}$ of $E$ 
(i.e., all the terms except for the first term in \eqref{Eompact})  
is   determined by 
$\begin{bmatrix}
\mathbf B^\top ,\mathbf C^\top,\mathbf D^\top,\mathbf  a^\top
\end{bmatrix}^\top\mathbf p = \mathbf\Gamma^\top \mathbf Q^\top\mathbf p =\mathbf  \Gamma^\top\mathbf u$,
which in turn is determined by a low dimensional variable $\mathbf u\triangleq  \mathbf Q^\top\mathbf p$.
The specific form of $E_{c}$ in terms of $\mathbf u$ is:
\begin{align}
 E_{c}( \mathbf u) = &
- \frac{1}{n_p} \|(\mathbf \Gamma^\top\mathbf u)_D \|^2  
+ \min_{s,\mathbf R} \{s^2 ( (\mathbf \Gamma^\top\mathbf u)_a  
- \frac{1}{n_p}    \|(\mathbf \Gamma^\top\mathbf u)_C \|^2)  \notag\\ 
& - 2s\, \text{tr}(\mathbf R [ \text{mat} ((\mathbf \Gamma^\top\mathbf u)_B)  
- \frac{1}{n_p} (\mathbf \Gamma^\top\mathbf u)_C (\mathbf \Gamma^\top\mathbf u)_D^\top  ]) \} 
\end{align}
Here  $(\mathbf \Gamma^\top\mathbf u)_{B}$ denotes the vector 
formed by the elements of vector $\mathbf \Gamma^\top\mathbf u$ 
with indices  equal to  row indices  of the
submatrix $\mathbf B$ in matrix $\begin{bmatrix}
\mathbf B^\top,\mathbf C^\top,\mathbf D^\top ,\mathbf  a^\top
\end{bmatrix}^\top$.
Vectors $(\mathbf \Gamma^\top\mathbf u)_{C}$, $(\mathbf \Gamma^\top\mathbf u)_{D}$ 
and $(\mathbf \Gamma^\top\mathbf u)_{ a}$ are similarly defined.
The dimension of $\mathbf u$ is $d^2+2d+1 $,
which is 
 independent of the cardinalities  of the two point sets.
This is the key reason why  our algorithm 
scales well with  problem size.

\subsubsection{2D case}
Although the preceding energy function derivation  directly applies to the 2D case,
an energy function with even fewer nonlinear terms
 can be derived for this case.
Assume the  rotation angle is $\beta$, 
then the rotation matrix is
$
 \mathbf R=\begin{bmatrix}
    \cos(\beta)& -\sin(\beta)\\ \sin(\beta) & \cos(\beta)
   \end{bmatrix}
$.
Let a unit vector $\mathbf r\triangleq [\cos(\beta),\sin(\beta)]^\top$,
then we have
\begin{gather}
\text{tr}(\mathbf R\mathbf H)=\mathbf r^\top \mathbf W \text{vec}(\mathbf H) \label{R_2D_relation}
\end{gather}
for any $2\times 2$ matrix $\mathbf H$,
where the constant matrix
 $\mathbf W\triangleq \begin{bmatrix}
            1 &0 &0& 1\\0 &1 &-1 &0
           \end{bmatrix}$.

Based on Eq. \eqref{R_2D_relation} and the fact that
$\max_{\mathbf r} \mathbf r^\top \boldsymbol\eta =\|\boldsymbol\eta\|$
for and 2D vector $\boldsymbol\eta$, 
we can rewrite the function $E(\mathbf P)$ in Eq. \eqref{E_P} as:
\begin{align}
 E(\mathbf P)=&
\mathbf 1^\top\mathbf P\mathbf { \widetilde y} 
- \frac{1}{n_p} \|\mathbf Y^\top\mathbf P^\top\mathbf  1\|^2  
+ \min_{s} \{s^2 (\mathbf {\widetilde x}^\top\mathbf P\mathbf  1   - \frac{1}{n_p}    \| \mathbf X^\top\mathbf P\mathbf  1\|^2)  \notag\\
& - 2s \| \mathbf W \text{vec}(\mathbf X^\top\mathbf P \mathbf Y - \frac{1}{n_p} \mathbf X^\top\mathbf P\mathbf 1 \mathbf 1^\top\mathbf P\mathbf Y)\|  \}
\end{align}

To facilitate the optimization of $E(\mathbf P)$,
$E(\mathbf P)$ needs to be expressed  in terms of vector $\mathbf p$.
Instead of using the  matrix denotation $\mathbf B$ as given in Eq. \eqref{B_denote},
we let
\[
\mathbf W\text{vec}(\mathbf X^\top\mathbf P\mathbf Y)\triangleq \mathbf  {\widetilde B} \mathbf p
\]
Based on the fact 
$\text{vec}(\mathbf M_1\mathbf  M_2\mathbf  M_3)= (\mathbf M_1\otimes\mathbf M_3^\top)\text{vec}(\mathbf M_2)$,
we have matrix
$
\mathbf {\widetilde B}=\mathbf W(\mathbf X^\top\otimes \mathbf Y^\top)
$.
Then, we can write  $E(\mathbf P)$ in terms of vector $\mathbf p$ as:
  \begin{align}
E(\mathbf p)= & 
(\mathbf 1_m^\top\otimes\mathbf { \widetilde y}^\top)\mathbf  p 
- \frac{1}{n_p} \|\mathbf D\mathbf  p \|^2  
+ \min_{s} \{s^2 (\mathbf a^\top\mathbf  p    - \frac{1}{n_p}    \|\mathbf C\mathbf  p \|^2)  \notag\\
& - 2s \|\mathbf {\widetilde B}\mathbf  p   - \frac{1}{n_p} \mathbf W\text{vec}(\mathbf  C\mathbf p\mathbf  p^\top \mathbf D^\top )\| \} \label{Eompact_2d}
\end{align}                  
Let the QR factorization of matrix $
\begin{bmatrix}
\mathbf {\widetilde B}^\top,\mathbf C^\top,\mathbf D^\top,\mathbf  a^\top
\end{bmatrix}
$ be
$
\mathbf Q\mathbf \Gamma=\begin{bmatrix}
\mathbf {\widetilde B}^\top ,\mathbf C^\top,\mathbf D^\top,\mathbf  a^\top
\end{bmatrix}
$.
In view of the form of $E$ in Eq. \eqref{Eompact_2d},
we can see that
the concave part $E_c$ of $E$
(i.e., all the terms except for the first term in \eqref{Eompact_2d})
is   determined by 
$\begin{bmatrix}
\mathbf{\widetilde B}^\top ,\mathbf C^\top,\mathbf D^\top,\mathbf  a^\top
\end{bmatrix}^\top\mathbf p =\mathbf \Gamma^\top \mathbf Q^\top\mathbf p =\mathbf \Gamma^\top\mathbf u$,
which in turn is determined by variable $\mathbf u\triangleq  \mathbf Q^\top\mathbf p$. 
The specific form of $E_c$ in terms of  $\mathbf u$ is
\begin{align}
 E_c(\mathbf u) =& 
- \frac{1}{n_p} \|(\mathbf\Gamma^\top\mathbf u)_D \|^2  
 + \min_{s} \{s^2 [ (\mathbf\Gamma^\top\mathbf u)_a  
- \frac{1}{n_p}    \|(\mathbf\Gamma^\top\mathbf u)_C \|^2 ]  \notag\\ 
& - 2s \|  (\mathbf\Gamma^\top\mathbf u)_{\widetilde B}  
- \frac{1}{n_p} \mathbf W \text{vec}((\mathbf\Gamma^\top\mathbf u)_C (\mathbf\Gamma^\top\mathbf u)_D^\top)  \| \} 
\end{align}
Here the vector $(\mathbf\Gamma^\top\mathbf u)_{\widetilde B}$ is similarly defined as vectors
$(\mathbf\Gamma^\top\mathbf u)_{C}$, $(\mathbf\Gamma^\top\mathbf u)_{D}$ and $(\mathbf\Gamma^\top\mathbf u)_{ a}$.

It is easy to verify that the dimension of
$\mathbf u$ is $3d+1|_{d=2}=7$,
which is lower than the dimension
$d^2+2d+1|_{d=2}=9$
as a result of directly applying the energy function derivation prior to this subsection to the 2D case.

\section{Optimization \label{sect:optimize}}


Our analysis in the previous section indicates that 
the nonlinear part of  $E(\mathbf p)$ is  determined by a low dimensional variable $\mathbf u$ and 
 is also concave.
Therefore it is natural to use the normal simplicial algorithm \cite{book_concave}, 
a BnB algorithm specifically designed for concave functions, to optimize $E$.

\subsection{Initial enclosing region \label{subsec:init}
}
\begin{figure}[h]
\centering
 \includegraphics[width=0.3\linewidth]{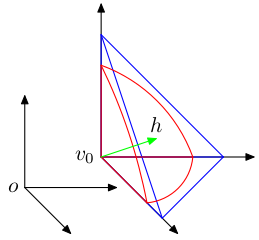}
 
\caption{
Red region: 
the intersection of  $U$ with an orthant of the new coordinate system.
Blue region: 
an enclosing simplex,
one of whose  faces supports $U$ 
and has a normal  $\mathbf h$ (in green)
 equal to  the normalized mean of the orthant axes. 
\label{bound_simp}}
\end{figure}

In the normal simplicial algorithm, 
simplexes are used to construct the convex envelopes of a concave function. 
Therefore the initial enclosing region should be chosen as a simplex or a collection of simplexes.
We use a collection of simplexes to enclose the feasible region 
$U\triangleq\{\mathbf u|\mathbf u=\mathbf Q^\top\mathbf p, \mathbf p\in \Omega\}$ 
as the resulting enclosing could be more tight,
where $\Omega$ denotes the feasible region of $\mathbf p$,
as determined by \eqref{k_card_P_const}.
The procedure is as follows. 
We first choose an interior point 
$\mathbf v_0=\mathbf Q^\top  \frac{n_p}{mn}\mathbf 1_{mn}$ of $U$,
which corresponds to the fuzziest point correspondence. 
We then construct a new coordinate system by translating  the coordinate system of $\mathbf u$
so that the new origin  locates at $\mathbf v_0$,
as illustrated in Fig. \ref{bound_simp}.
We now construct each enclosing simplex 
as the intersection of an orthant of the new coordinate system 
with a half space containing $U$, 
whose face 
supports $U$ and has a normal vector $\mathbf h$ chosen as the normalized mean of the orthant axes,
as illustrated in 
Fig. \ref{bound_simp}.

The distance from $\mathbf v_0$ to the supporting plane with normal $\mathbf h$ can be computed as: 
\begin{equation}\label{support_plane}
 \max \{ \mathbf h^\top (\mathbf Q^\top \mathbf p - \mathbf v_0) | \mathbf p\in \Omega\}
\end{equation}
This is a k-cardinality linear assignment problem 
which can be either directly solved \cite{k_card_assign_direct} 
or transformed into a standard linear assignment problem \cite{k_card_assign_transform}
(we adopt the latter approach and choose the Jonker-Volgenant algorithm \cite{LAPJV}
for the resulting problem in this paper). 
The supporting plane with normal $\mathbf h$ can then be completely determined.
In turn, the vertices of the enclosing simplex can be recovered which has $\mathbf v_0$ as one of its vertices.


\subsection{Choice of  $\mathbf H$ in approach one of our algorithm\label{subsec:H}}
For approach one of our algorithm, we need to ensure $E$ is concave over all the enclosing simplexes.
Based on proposition 1,
it suffices if $\mathbf A(\mathbf u)+\mathbf H\succ0$ for any $\mathbf u$ belonging to  the enclosing simplexes.
This condition can be satisfied by setting
the eigenvalues  of $\mathbf H$ to be large enough.
The procedure is as follows.
Assume  the eigenvalues of  $\mathbf A(\mathbf v_i)$ are $\lambda^j_i$,
where $\mathbf v_i$ is a vertex of  the enclosing simplexes.
We choose a scalar   
$\lambda_0 =\min \{\min_{i,j} {\lambda_i^j},0 \}$
and   set $\mathbf H=(-\lambda_0+\epsilon_0)\mathbf  I$,
where $\epsilon_0$ is a  small positive value 
(we set $\epsilon_0=10^{-5}$ in this paper).
We now have
$\mathbf A(\mathbf v_i)
+\mathbf H\succ 0$ for any vertex $\mathbf v_i$ of the enclosing simplexes.
Since 
$\mathbf A(\mathbf u)+\mathbf H\succ 0 $ is a spectrahedra and thus  convex as indicated by proposition 1,
we therefore have  $\mathbf A(\mathbf u) + \mathbf H\succ 0$ for any $\mathbf u$
belonging to the enclosing simplexes.

\subsection{Lower bounds}

The convex envelope $E_{cs}$ of the concave part $E_c(\mathbf u)$ of  $E$ 
over a simplex $S=[\mathbf v_1,\ldots,\mathbf v_{n_u +1}]$ is
the unique affine function which coincides with $E_c$ 
at the vertices $\mathbf v_1,\ldots,\mathbf v_{n_u+1}$  \cite{book_concave},
i.e.,
$
E_{cs}(\mathbf u)= \sum_{i=1}^{n_u+1} \alpha_i E_c(\mathbf v_i) 
$
with 
$\mathbf u=\sum_{i=1}^{n_u+1} \alpha_i \mathbf v_i$,
$\sum_{i=1}^{n_u+1} \alpha_i=1$,
$\alpha_i\ge 0$,
$\forall i$.
Here $n_u$ denotes the dimension of $\mathbf u$.
Based on this result,
the lower bound of $E$ for region $\Omega\cap S$ can be obtained as the optimal value of  the following linear program:
\begin{align}
& \min_{\alpha_i,\mathbf p}\  \sum_{i=1}^{n_u+1} \alpha_i E_c(\mathbf v_i) 
 +(\mathbf 1_m^\top\otimes\mathbf{\widetilde y}^\top)\mathbf  p \notag\\
& s.t. \sum_{i=1}^{n_u+1} \alpha_i\mathbf v_i=\mathbf Q^\top\mathbf p,
 \sum_{i=1}^{n_u+1} \alpha_i=1,\alpha_i\ge 0,\forall i,\mathbf p\in \Omega \label{lower_bound_prob}
\end{align}
By tweaking this linear program, 
in Sec. \ref{subsec:fast_bound},
we will propose an alternative lower bounding problem  
which is much more efficient to solve.

\subsubsection{Value of $E( \mathbf v_i)$ in approach two of our algorithm}
For approach two of our algorithm, 
the value of  $E( \mathbf v_i)$  needs  to be determined.
In 3D case, $E( \mathbf v_i)$ has the following form:
\begin{align}
 E(\mathbf v_i)= &
- \frac{1}{n_p} \|(\mathbf \Gamma^\top\mathbf v_i)_D \|^2  
+ \min_{s,\mathbf R} \{s^2 [ (\mathbf \Gamma^\top\mathbf v_i)_a  
- \frac{1}{n_p}    \|(\mathbf \Gamma^\top\mathbf v_i)_C \|^2 ]  \notag\\ 
& - 2s\, \text{tr}(\mathbf R [ \text{mat} ((\mathbf \Gamma^\top\mathbf v_i)_B) 
- \frac{1}{n_p} (\mathbf \Gamma^\top\mathbf v_i)_C (\mathbf \Gamma^\top\mathbf v_i)_D^\top  ]) \} 
\label{value_E_v}
\end{align}
Let matrix
\[
\mathbf G\triangleq   \text{mat} ((\mathbf \Gamma^\top\mathbf v_i)_B) 
- \frac{1}{n_p} (\mathbf \Gamma^\top\mathbf v_i)_C (\mathbf \Gamma^\top\mathbf v_i)_D^\top  
\]
and let $\mathbf U\mathbf S\mathbf V^\top$ be the singular value decomposition of $\mathbf G^\top$,
where $\mathbf S$ is a diagonal matrix and
the columns of  $\mathbf U$ and $\mathbf V$ are orthogonal unity vectors.
Then the optimal rotation matrix $\mathbf R$ solving problem  \eqref{value_E_v} is
$ \mathbf R^*=\mathbf U\text{diag}(\begin{bmatrix}
                     1,\ldots,1,\det(\mathbf U\mathbf V^\top)
                    \end{bmatrix})\mathbf V^\top$ \cite{CPD}.
Here $\text{diag}(\cdot)$ denotes converting a vector into a diagonal matrix,
and $\det(\cdot)$ denotes the determinant of a square matrix.
By substituting $ \mathbf R^*$ back into \eqref{value_E_v}, $\mathbf R$ is eliminated and
we get a (possibly concave) quadratic program only  in one variable $s$.
If the range of $s$ is $\underline{s}\le s \le \overline{s}$,
then one can easily solve  this  quadratic program 
by comparing the function values at the two boundary points $\underline{s}$, $\overline{s}$ and 
the extreme point to obtain  the optimal $s$.

\textbf{2D case:}
For 2D case, we have
\begin{align}
 E(\mathbf  v_i)= &
- \frac{1}{n_p} \|(\mathbf \Gamma^\top\mathbf v_i)_D \|^2  
+ \min_{s} \{s^2 ( (\mathbf \Gamma^\top\mathbf v_i)_a  
- \frac{1}{n_p}    \|(\mathbf \Gamma^\top\mathbf v_i)_C \|^2)  \notag\\ 
& - 2s \|  (\mathbf \Gamma^\top\mathbf v_i)_{B_2}  
- \frac{1}{n_p} \mathbf W \text{vec}((\mathbf \Gamma^\top\mathbf v_i)_C (\mathbf \Gamma^\top\mathbf v_i)_D^\top)  \| \} 
\end{align}
This is a quadratic program in only one variable  $s$.
Hence the optimal $s$ can  similarly be solved based on the above discussion.

\subsection{Division of a simplex\label{subsec:branch}}
Since the BnB algorithm is used for optimization, 
during the branching phase, 
a chosen simplex needs to be subdivided into several smaller simplexes.
We adopt the following simple strategy to divide a simplex. For a chosen simplex, 
the longest edge is bisected. 
This results in two sub-simplexes. 
It has been proved that such a subdivision scheme leads to a BnB algorithm that converges \cite{book_concave}.

\subsection{The normal simplicial BnB algorithm}

With the preparation from the previous subsections, 
We are now ready to describe the algorithm for minimizing $E(\mathbf p)$. 
During initialization, 
a set of simplexes whose union contains the feasible region $\Omega$ is computed. 
Then in each iteration, 
the simplex yielding the lowest lower bound among all the simplexes is further subdivided 
so as to improve the lower bound of $E$ for $\Omega$.
Meanwhile, 
the upper bound is updated by evaluating $E$ 
with solutions of the linear programs used to compute the lower bounds. 
The pseudo-code of the algorithm is summarized in Algorithm \ref{normal_sim_algo}.

\begin{algorithm}

\textbf{Initialization}

Select tolerance error  $\epsilon>0$. 

Find a collection of simplexes $\{S_i\}$ such that $U \subset {\cup_i} S_i$
according to Sec. \ref{subsec:init}.
(For approach one,
choose  $\mathbf H$ according to Sec. \ref{subsec:H}.)
Set $\mathscr M_1=\mathscr N_1=\{S_i\}$,
where $\mathscr M_1$ denotes the collection of all simplexes and 
$\mathscr N_1$ denotes the collection of active simplexes.


\For{$k=1,2,\ldots$}{

For each simplex $S\in\mathscr N_k$,
solve the linear program \eqref{lower_bound_prob}
to obtain a basic optimal solution $\boldsymbol\omega(S)$ and the optimal value $\beta(S)$.
$\beta(S)$  is the lower bound of $E$ for region $\Omega\cap S$.

Let $\mathbf p^k$ equal  the best  among all feasible solutions so far encountered:
$\mathbf p^{k-1}$ and all $\boldsymbol\omega(S),S\in \mathscr N_k$.
Delete all simplexes $S\in \mathscr M_k$ such that $\beta(S)\ge E(\mathbf p^k)-\epsilon$.
Let $\mathscr R_k$ be the remaining collection of simplexes.

If $\mathscr R_k=\emptyset$, terminate: $\mathbf p^k$ is the global $\epsilon$-minimal solution. 
Otherwise, go to the next step.

Select the simplex to be divided: $S_k\in\arg\min\{\beta(S)|S\in\mathscr R_k\}$.

Divide $S_k$ according to Sec. \ref{subsec:branch}
to get two sub-simplexes $S_{k1}$ and $S_{k2}$.

Let $\mathscr N_{k+1}=\{S_{k1},S_{k2}\}$ and  
$\mathscr M_{k+1}=(\mathscr R_k\backslash \{S_k\})\cup\mathscr N_{k+1}$.
}
\caption{The normal simplicial algorithm  for minimizing $E$ \label{normal_sim_algo}}

\end{algorithm}


\subsection{A new fast lower bounding scheme\label{subsec:fast_bound}}
Our algorithm is an instance of the BnB technique, 
therefore it contains three basic subroutines: branching, finding upper and lower bounds. 
It is obvious that the lower bounding subroutine \eqref{lower_bound_prob} 
requires much more time to compute than the other two subroutines 
since it is a generic linear program, 
for which there are no efficient algorithms. 
To address this problem, in this subsection,
we will propose an alternative lower bounding scheme which is more efficient to compute.

To this end,
a natural idea is to drop the inequality constraints $\alpha_i\ge0,\forall i$ in \eqref{lower_bound_prob},
then there are only linear equality constraints on $\alpha_i$.
Therefore  $\alpha_i$ can be eliminated via algebraic substitution
 and we  arrive at the following equivalent problem:
\begin{align}
\min  E_S (\mathbf p) =&( [ E_c(\mathbf v_1),\ldots,E_c(\mathbf v_{n_u})]
- E_c(\mathbf v_{n_u+1}) \mathbf 1_{n_u}^\top) 
( [\mathbf v_1,\ldots,\mathbf v_{n_u}] -\mathbf v_{n_u+1} \mathbf 1_{n_u}^\top)^{-1}   
 (\mathbf Q^\top\mathbf  p-\mathbf v_{n_u+1}) \notag\\
& + E_c(\mathbf v_{n_u+1})  +(\mathbf 1_m^\top\otimes\mathbf {\widetilde y}^\top)\mathbf  p \notag\\
s.t. \quad  \mathbf   p\in \Omega \label{lower_bound_fast} & 
\end{align}
Problem \eqref{lower_bound_fast} is a k-cardinality linear assignment problem
which can be efficiently solved by the combinatorial optimization algorithms
mentioned in Sec. \ref{subsec:init}.
Note that simplex $S$ will not degenerate throughout the BnB iterations
and therefore matrix 
$[\mathbf v_1,\ldots,\mathbf v_{n_u}] -\mathbf v_{n_u+1} \mathbf 1_{n_u}^\top$ is always invertible.
We have the following proposition.
\begin{proposition}
The optimal value of problem \eqref{lower_bound_fast} is a lower bound of $E$ for region $\Omega\cap S$.
\end{proposition}
{\proof
Problem \eqref{lower_bound_fast} is a relaxed version  of \eqref{lower_bound_prob} 
by dropping the constraints $\alpha_i\ge 0,\forall i$.
Therefore  the optimal  value of \eqref{lower_bound_fast} will not be greater than that of \eqref{lower_bound_prob},
whereas  solving \eqref{lower_bound_prob} yields a lower bound of $E$ for region $\Omega\cap S$.
}

\begin{figure}[h]
\centering
\subfigure{\includegraphics[width=0.38\linewidth]{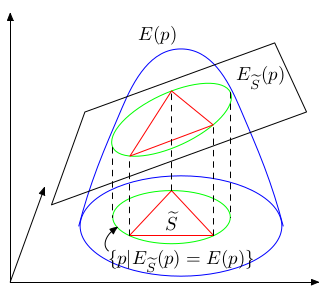}}
\caption{Ellipse-like closed curve $\{\mathbf p|  E_{\widetilde S}(\mathbf p)= E(\mathbf p) \}$ 
separates the solution space into two disjoint regions.
Simplex $\widetilde S$ is contained in the region  $\{\mathbf p|  E_{\widetilde S}(\mathbf p)\le E(\mathbf p) \}$.
\label{ellipse}}
\end{figure}

What remains  is to check whether  the  lower bound computed by  \eqref{lower_bound_fast} is 
close to the original lower bound.
In iteration $k$ of our algorithm,
the lowest one among all the lower bounds corresponding to simplexes in $\mathscr M_k$ 
is chosen as the lower bound of $E$ for the feasible region $\Omega$.
Therefore only the lowest  lower bound determines the quality of a bounding scheme.
Without loss of generality,
let us assume that  $\widetilde S$ is the simplex yielding the lowest  lower bound 
when using  \eqref{lower_bound_fast}  to compute the lower bound of $E$ for region $\Omega$.
It's apparent that there are two possibilities 
for the location of the optimal solution $\mathbf p^*$ of problem \eqref{lower_bound_fast}: 
either $\mathbf p^*\in\{\mathbf p|  E_{\widetilde S}(\mathbf p)\le E(\mathbf p) \}$ or 
$\mathbf p^*\in\{\mathbf p| E_{\widetilde S}(\mathbf p)> E(\mathbf p)\}$,
as illustrated in Fig. \ref{ellipse}.
Note that the latter case is impossible
since in this case, 
$E_{\widetilde S}(\mathbf p^*)$ will be strictly larger 
than the minimum  value of $E$ over  $\Omega$,
violating  the assumption that $E_{\widetilde S}(\mathbf p^*)$ is a lower bound of $E$ for region $\Omega$.
Therefore it can only happen that $\mathbf p^*\in\{\mathbf p|  E_{\widetilde S}(\mathbf p)\le E(\mathbf p) \}$.
Since $p^*$ can only be obtained  at one of the vertices of $\Omega$.
This also indicates that  $\{\mathbf p|  E_{\widetilde S}(\mathbf p)\le E(\mathbf p) \}$
contains a segment of the boundary of $\Omega$.

From Fig. \ref{ellipse}, we can see that $\{\mathbf p|  E_{\widetilde S}(\mathbf p)\le E(\mathbf p) \}$
 is an ellipsoid-like region containing  and circumscribing the simplex $\widetilde S$. 
Therefore,
under the condition that the gradient of the plane $E_{\widetilde S}$ is not large,
the  lower bounds computed via \eqref{lower_bound_prob} and  via \eqref{lower_bound_fast}
will be close to each other.



\subsubsection{Effectiveness of the fast bounding scheme\label{new_bound_effect}}
\begin{figure}[t]
\includegraphics[width=1\linewidth]{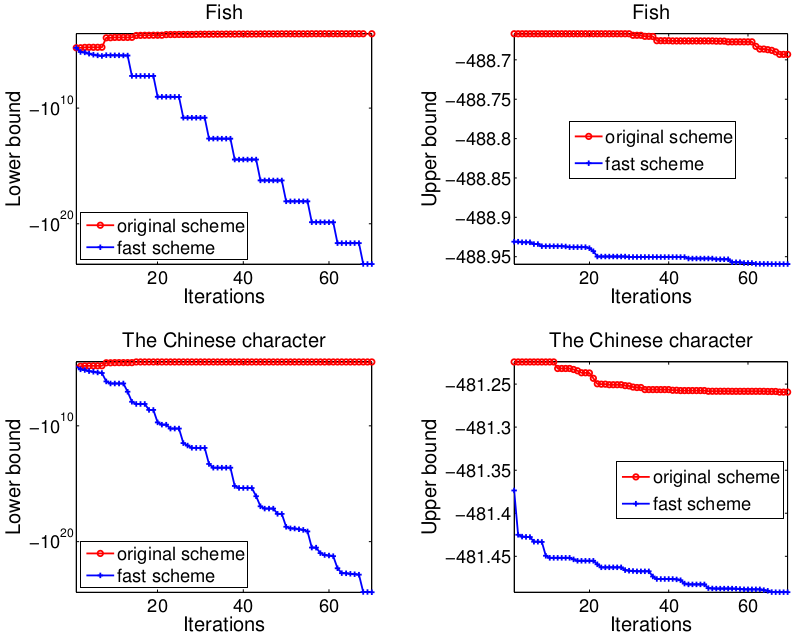}
%
%

\caption{
Mean of the lower bounds (left column) and upper bounds (right column) for the feasible region by the two bounding schemes
in each iteration of our algorithm.
The outlier test in section \ref{subsec:regu2Dtest}  is used
and 2D similarity transformation is chosen for our algorithm.
%
%
\label{two_bounds_compare}
}
\end{figure}

In this subsection,
we will compare the performances of our algorithm under
either the original bounding scheme or the fast bounding scheme.
We first  use  the outlier test  as described in section \ref{subsec:regu2Dtest} 
(which tests our first algorithm using regularization on transformation parameters)
for comparison
where the outlier to data ratio is chosen as $0.5$.
The lower and upper bounds for the feasible region 
generated by the two bounding schemes in each iteration of
our algorithm are shown in
figure \ref{two_bounds_compare}.
It can be seen that the difference between the lower bounds  generated by the two bounding schemes widens
as the number of iterations increases.
The main reason is that 
the gradient of the energy function $E$ is large at the boundary of the feasible region of $\mathbf u$
(since $\mathbf H$ is chosen so that $\mathbf A(\mathbf u)+\mathbf H$ is barely positive definite
within the feasible region of $\mathbf u$,
 the value of $E$ (which contains inversion of $\mathbf A(\mathbf u)+\mathbf H$)
will  change dramatically at the boundary of the feasible region of $\mathbf u$).
Consequently,
the gradient of the lower bounding plane $E_{\widetilde S}$ is also large
(particularly when the simplex ${\widetilde S}$  becomes small)
since $\{\mathbf p|  E_{\widetilde S}(\mathbf p)\le E(\mathbf p) \}$
contains a segment of the boundary of the feasible region of $\mathbf u$.
As a result, 
the minimum value of $E_{\widetilde S}$ calculated within the simplex ${\widetilde S}$ 
(which is the lower bound by the original scheme)
will differ significantly from the minimum value of $E_{\widetilde S}$ calculated within the ellipsoid-like region
$\{\mathbf p|  E_{\widetilde S}(\mathbf p)\le E(\mathbf p) \}$
(which is the lower bound by the fast scheme).
Nevertheless,
from the right column of  figure  \ref{two_bounds_compare}, 
one can see that the upper bound generated by the fast bounding scheme is always lower than that generated by the original
scheme,
whereas the solution of the algorithm is chosen based on the upper bound.
Therefore, this indicates that the proposed fast bounding scheme is better at locating good solutions than the original scheme.
Figure  \ref{two_bounds_compare}  also suggests that for fast bounding scheme,
since the difference between upper and lower bounds never shrinks to zero,
we cannot use the difference between the upper and lower bounds (i.e., tolerance error)
to determine when to terminate our algorithm.
Instead, in this paper, 
we will use the maximum search depth (chosen as 15) of the BnB algorithm as the termination criterion.
However,
this decision will causes our algorithm not to be $\epsilon-$globally optimal.
Nevertheless,
we empirically found that our method with this decision performs very well in practice.

The average run time of our algorithm under different choices of bounding schemes are listed in Table
\ref{bounding_methods_time}. 
It can be seen that the speed of our algorithm using the fast bounding scheme is $254$ times
that using the original bounding scheme, 
demonstrating its high computational efficiency.

\begin{table}[h]
\centering
\caption{
Average run time  of our algorithm under different choices of bounding schemes (in seconds).}
\label{bounding_methods_time}
 \begin{tabular}{|c|c|c|}
\hline
              &fish &the Chinese character \\\hline
original scheme &  722.5290  & 1061.5   \\\hline
fast scheme     & \textbf{2.7529} &\textbf{4.2801}    \\\hline
\end{tabular} 
\end{table}

We then use the outlier test  as described in section \ref{subsec:simi2Dtest} 
(which tests our second algorithm using constraints on transformation parameters)
for comparison,
where the outlier to data ratio is chosen as $0.5$.

\begin{figure}[ht]

\includegraphics[width=\linewidth]{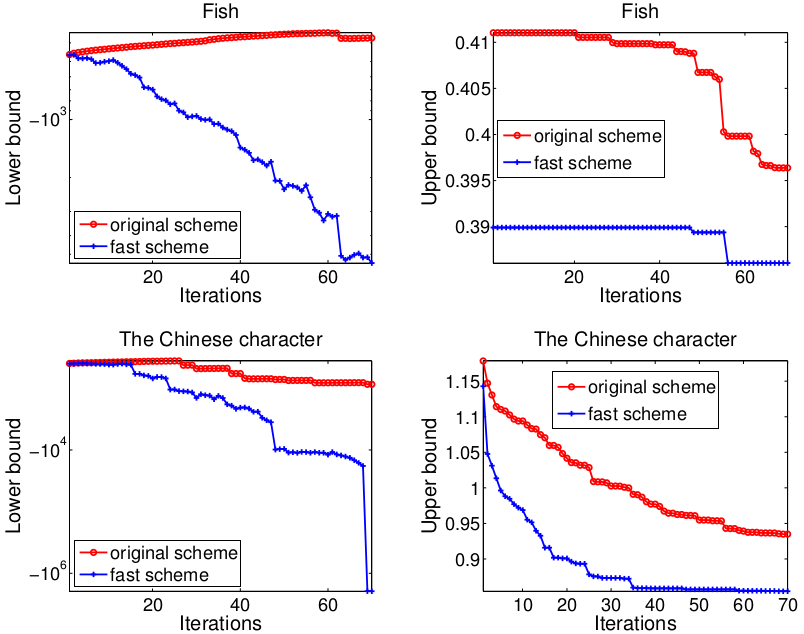}
%

\caption{
Mean of the lower bounds (left column) and upper bounds (right column) for the feasible region by the two bounding schemes
in each iteration of our algorithm.
The outlier test in section \ref{subsec:simi2Dtest}  is used
and 2D similarity transformation is chosen for our algorithm.
%
%
\label{two_bounds_compare_alg2}
}
\end{figure}

The lower and upper bounds for the feasible region 
by the two  schemes in each iteration of
our algorithm are shown in  figure \ref{two_bounds_compare_alg2}.
It can be seen that the difference between the lower bounds  generated by the two bounding schemes widens
as the number of iterations increases,
albeit not as quickly  as in the previous test.
Similar reason as in the previous test can be said 
about this phenomenon.

\subsection{GPU speed-up}
Our algorithm  initially needs to compute $2^{n_u}$ enclosing simplexes (see section \ref{subsec:init})
which corresponds to $2^{n_u}$ orthants of the space of $\mathbf u$
by solving the linear assignment problem \eqref{support_plane}.
Then, in step 5 of  algorithm 1,
the linear assignment problem \eqref{lower_bound_fast} needed to be solved for a set $\mathscr N_k$ of simplexes.
Initially, the size of $\mathscr N_1$ is  $2^{n_u}$.
When $n_u$ is large
(e.g., in the case of 3D similarity registration),
the above routines cost considerable amount of time.
Therefore, 
it's desirable that the above routines can be computed as fast as possible.
Fortunately,
note that the above routines are  independent repetitive routines,
Therefore,
it's ideal for them to be implemented in parallel.
In this paper,
we implement the above routines by using the parallel programming toolbox provided by Matlab
on an Nvidia Quadro K2200 GPU  card.
Our experimental results show that doing so can bring about 4x speed-up improvement
compared with pure CPU implementation.

\section{Experimental results\label{sec:exp}}

We implement our method under the Matlab 2014a environment
and compare it with other methods  on a PC with 2.4 GHz CPU and 8G RAM.
For  the methods to be compared which only output point correspondences,
we use the correspondences generated  by the methods 
to find the best affine transformations between  two point sets.
We define error  as  mean of the Euclidean distances 
between the transformed ground truth model inliers and their corresponding scene inliers.


Since there are two versions of  our algorithm 
which have different requirements on values of transformation parameters,
accordingly,
we will conduct different experiments to respectively test their performances.

\subsection{Experiments on point sets with no rotation between them }
In this subsection,
we will test our first algorithm which uses regularization on transformation 
and thus does not allow arbitrary rotation between two point sets.
We compare it with RPM \cite{RPM_TPS}, CPD \cite{CPD} and MG \cite{kernel_Gaussian}, 
whose source codes are freely available. 
These methods represent state-of-the-arts, 
only utilize the point position information for matching, and 
are capable of handling partial overlaps between two point sets.

\subsubsection{2D synthesized datasets \label{subsec:regu2Dtest}}

Two choices of transformations are considered for our method:
2D similarity  and affine transformations.
For 2D similarity transformation,
we let ${\boldsymbol\theta}=\begin{bmatrix} \theta_1, \ldots , \theta_4 \end{bmatrix}^\top$.
Here $[\theta_3, \theta_4]^\top$ represents  translation 
and  $\theta_1=r\cos(\beta)$ and $\theta_2=r\sin(\beta)$,
with  $r$ denoting  scale and $\beta$ denoting rotation angle.
Then we  have the Jacobian matrix
$\mathbf J(\mathbf x_i)=\begin{bmatrix}x_i^1&- x_i^2&1&0\\ x_i^2&x_i^1&0&1\end{bmatrix}$.
It can be verified that the rows of   
$\mathbf B_2=\mathbf B(1,:)$ and $\mathbf D_2=\mathbf D([1,2],:)$ 
constitute the unique  rows of $\mathbf B$ and $\mathbf D$
not equal to multiple of $\mathbf 1_{mn}^\top$, respectively.

For 2D affine transformation,
we let
${\boldsymbol\theta}=\begin{bmatrix} \theta_1, \ldots, \theta_6 \end{bmatrix}^\top$
with $[\theta_1,\ldots, \theta_4 ]^\top$ 
being the parameters of the linear part of the transformation and
$[\theta_5, \theta_6]^\top$ representing  translation.
Then we  have
$\mathbf J(\mathbf x_i)=\begin{bmatrix}
x_i^1& x_i^2&0&0&1&0\\
0&0&x_i^1& x_i^2&0&1
\end{bmatrix}$.
It can be verified that the rows of 
$\mathbf B_2=\mathbf B([1,2,6],:)$ and $\mathbf D_2=\mathbf D([1,3],:)$ 
constitute the unique rows of $\mathbf B$ and $\mathbf D$ 
not equal to multiple of $\mathbf 1_{mn}^\top$, respectively.


Affine transformation is used for RPM and
rigid transformation is used for CPD and MG 
(other types of transformations are found to be far less robust 
for the types of experiments conducted in this paper).

\begin{figure}[ht]
\centering
\includegraphics[width=\linewidth]{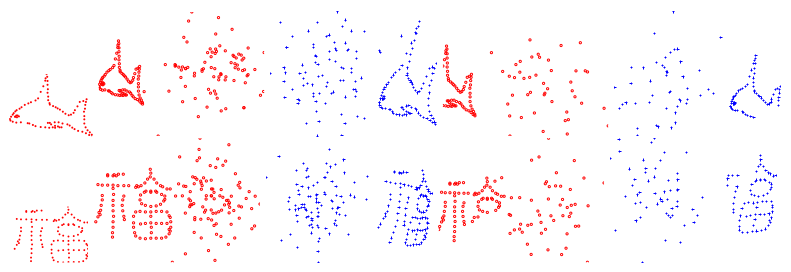}

%
\caption{
Left column: the prototype shapes.
For the remaining columns:
examples of model and scene point sets in the 
outlier (columns 2, 3) and occlusion + outlier (columns 4, 5) tests. 
\label{nonrot_2D_test_data_exa}}
\end{figure}

Two categories of tests are used to evaluate  performances of different methods:
1) \textbf{Outlier test}.
Equal number of normally distributed random outliers are added to 
different sides of the prototype shape to generate  two point sets
so as to simulate outlier disturbance, 
as illustrated in columns 2, 3 of Fig. \ref{nonrot_2D_test_data_exa}.
2) \textbf{Occlusion + Outlier test}.
First, 
equal degree of occlusions are applied to the prototype shape 
to generate  two point sets, respectively.
We simulate occlusion by first finding the shortest Hamiltonian circle of the prototype point set 
(via solving a traveling salesman problem) 
and then retaining a segment of the circle starting at a random point.
Then,
a fixed number of normally distributed random outliers (outlier to data ratio is fixed to 0.5) 
are added to different sides of the two point sets
so as to simulate outlier disturbance,
as illustrated in columns 4, 5 of Fig. \ref{nonrot_2D_test_data_exa}.
For all the above tests,
 random scaling within range  from $0.5$ to $1.5$ is applied to the prototype shape 
when generating the model point set and
a moderate amount of nonrigid deformation is applied to the prototype shape 
when generating the scene point set.
Two shapes \cite{RPM_TPS}, a fish and a character,
as shown in the left column of Fig. \ref{nonrot_2D_test_data_exa},
are used as the prototype shape, respectively.

\begin{figure}[ht]
\centering
\includegraphics[width=\linewidth]{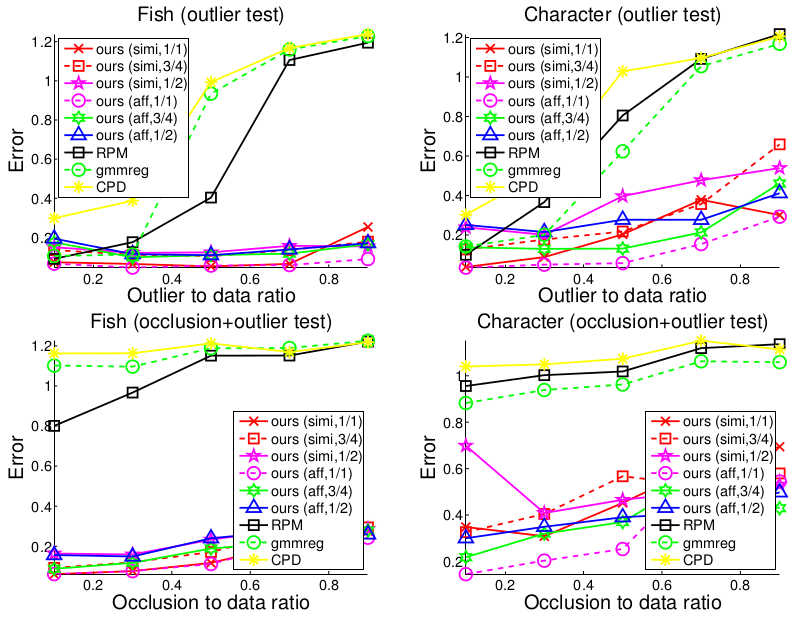}

\caption{
Average  matching errors by our method under different transformation choices (similarity and affine)
and with different $n_p$ values (chosen from $1/2$ to $1/1$ the ground truth value) and
other methods over 100 random trials
for the 2D outlier and occlusion+outlier tests.
\label{2D_nonrotate_sta}}
\end{figure}

\begin{figure}[ht]
\centering

\includegraphics[width=\linewidth]{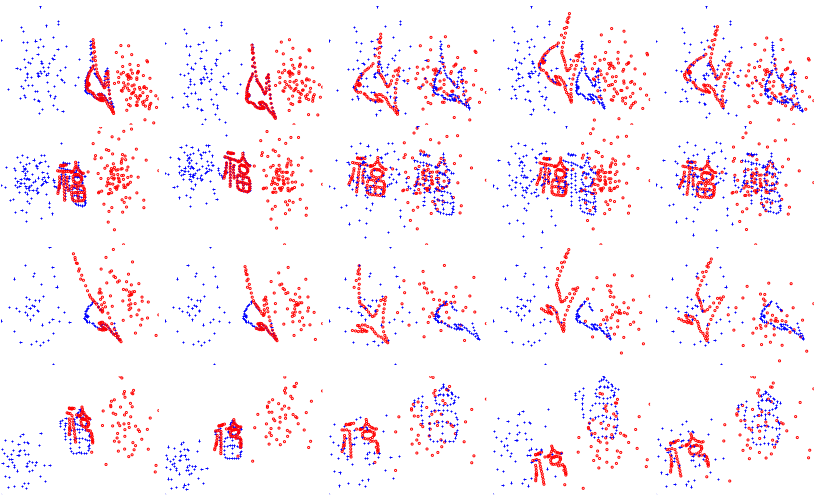}

    \caption{
Examples of matching results by different methods 
in the outlier (top 2 rows) and occlusion + outlier (bottom 2 rows) tests.
\label{two_test_exa}}
\end{figure}

The average matching errors of different methods
are shown in Fig. \ref{2D_nonrotate_sta}.
It can be seen that our method performs much better than other methods,
particularly in the occlusion+outlier test
where there is a large margin between the errors of our method and those of other methods.
This demonstrates our method's robustness to disturbances.
Among the different transformation choices of our method,
our method using affine transformation performs relatively better than our method using similarity transformation.
Among the different $n_p$ choices of our method,
our method with $n_p$ chosen close to the ground truth value performs relatively better than
our method with $n_p$ chosen far away from the ground truth value.
Examples of matching results by different methods are shown in Fig. \ref{two_test_exa}.

The average running times of our method using similarity or affine transformations,
RPM, gmmreg and CPD  are 
    3.3286, 
   26.6627,
      1.8421,
    0.0613 and
    0.0623 seconds, respectively.
It can be seen that our method using similarity transformation has similar running time as RPM.
Among the different transformation choices of our method,
our method using similarity transformation is almost an order-of-magnitude  faster than
our method using affine transformation.
This is because affine transformation have more parameters than similarity transformation,
which results in a higher dimensional search space in our method
and thus the BnB algorithm needs more time  to converge.


\subsubsection{3D synthesized datasets \label{subsec:regu3Dtest}}   
Since 3D affine transformation contains too many parameters
which causes our method to converge too slowly,
this  transformation will not be tested for our method.
Instead, we consider a 3D transformation  consisting of nonuniform scaling  and translation for our method:
$T( \mathbf x_i| \boldsymbol\theta)= \begin{bmatrix}
  \theta_1 x_i^1  +\theta_4, &   \theta_2 x_i^2  +\theta_5, &
    \theta_3  x_i^3  +\theta_6
         \end{bmatrix}^\top
$ 
with $\boldsymbol\theta=[\theta_1,\ldots,\theta_6]$.
We have the Jacobian matrix 
$\mathbf J(\mathbf x_i)=\begin{bmatrix}
           x_i^1 & 0 & 0 & 1 & 0 & 0\\
  0 &  x_i^2 & 0 & 0 & 1 & 0\\
  0 & 0 &  x_i^3  & 0 & 0 & 1
        \end{bmatrix}
$.
It can be verified that the rows of 
$\mathbf B_2=\mathbf B([1,5,9],:)$  and $\mathbf D_2=\mathbf D([1,5,9],:)$ 
constitute the unique rows of $\mathbf B$ and  $\mathbf D$
not equal to multiple of $\mathbf 1_{mn}^\top$, respectively.

\begin{figure}[ht]

\includegraphics[width=1\linewidth]{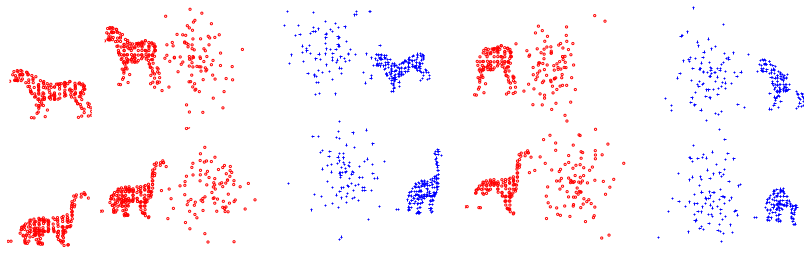}

%
\caption{
Left column: the prototype shapes. 
For the remaining columns: 
examples of model and scene point sets in the  outlier (columns 2, 3)
and  occlusion+outlier (columns 4, 5) tests.
\label{nonrot_3D_test_data_exa}}
\end{figure}

Analogous  to the experimental setup in  the preceding subsection,
we use two categories  of tests to evaluate  performances of different methods:
1) \textbf{Outlier test} and
2) \textbf{Occlusion + Outlier test},
as illustrated in  Fig. \ref{nonrot_3D_test_data_exa}.
Two shapes\footnote{These shapes can be downloaded at 
the  AIM@SHAPE Shape Repository: \textit{http://shapes.aimatshape.net/}.}, a horse  and a  dinosaur, 
as shown in the left column of Fig.  \ref{nonrot_3D_test_data_exa},
are used as the prototype shape, respectively.

\begin{figure}[ht]
\centering

\includegraphics[width=1\linewidth]{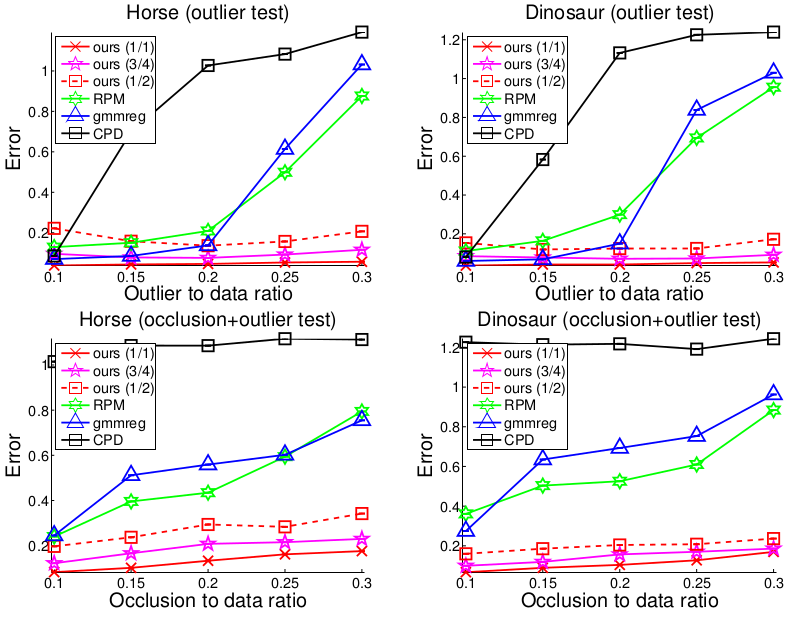}

\caption{Average  matching errors by our method with
different $n_p$ values (chosen from $1/2$ to $1/1$ the ground truth value)
and other  methods over 100 random trials
for the 3D outlier and occlusion+outlier tests.
\label{3D_nonrotate_sta}}
\end{figure}

To make a fair comparison,
the same 3D transformation of nonuniform scaling + translation is employed by all the comparison methods.
The average matching errors of different methods are shown in Fig. \ref{3D_nonrotate_sta}.
It can be seen that our method performs much better than other methods
and its errors keep almost unchanged with the increase of severity  of disturbances.
This demonstrates our method's strong robustness to disturbances.
Examples of matching results by different methods are shown in Fig. \ref{3D_nonrotate_exa}.

\begin{figure}[ht]

\includegraphics[width=1\linewidth]{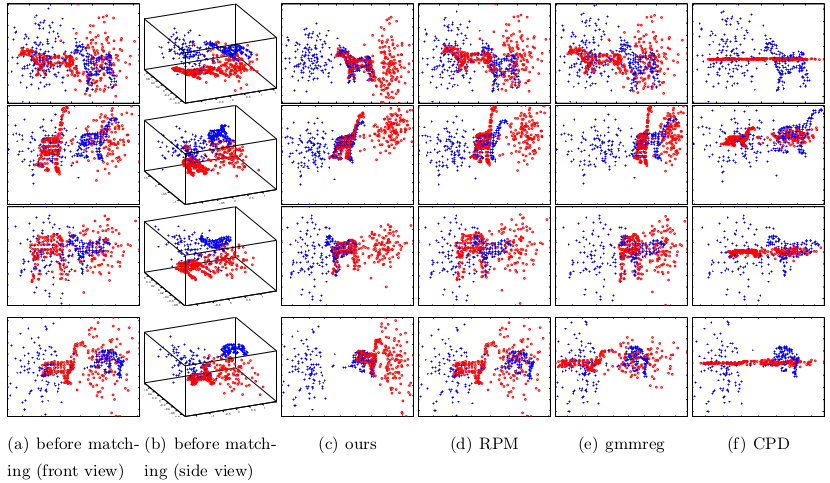}

  \caption{
Examples of   matching results by different methods in the outlier (top 2 rows) 
and  occlusion+outlier (bottom 2 rows) tests.
  \label{3D_nonrotate_exa}}
    \end{figure}

The average running times of our method,
RPM, gmmreg and CPD  are 
19.8839, 
     2.5296,
    0.2842 and
    0.0957 seconds, respectively.

\subsection{Experiments on point sets with random rotations  between them}
In this subsection,
we will test our second algorithm which uses constraints on transformation
and allows arbitrary rotation and uniform scaling within a range between two point sets.
We compare it with Go-ICP \cite{Go-ICP}
which is globally optimal,
only utilizes point position information and 
allows arbitrary rotations between two point sets.
The source code of Go-ICP is generously provided by the author. 
The range of scale $s$ in our method is set as $0.5\le s\le 1.5$.

\subsubsection{2D synthesized datasets  \label{subsec:simi2Dtest}}

\begin{figure}[ht]
\centering

\includegraphics[width=1\linewidth]{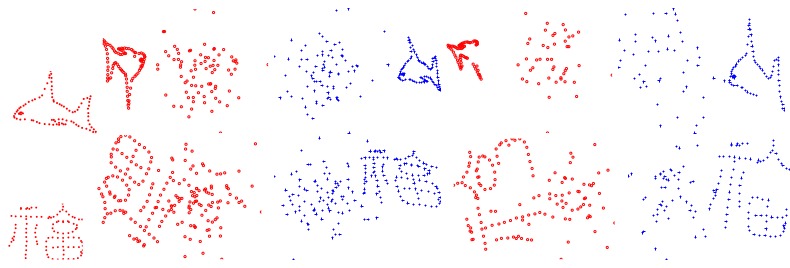}

%
\caption{
Left column: the prototype shapes.
For the remaining columns:
examples of model and scene point sets in the 
outlier (columns 2, 3) and occlusion + outlier (columns 4, 5) tests. 
\label{rot_2D_test_data_exa}}
\end{figure}

Analogous  to the experimental setup in   subsection \ref{subsec:regu2Dtest},
we use two categories  of tests to evaluate  performances of different methods:
1) \textbf{Outlier test} and
2) \textbf{Occlusion + Outlier test},
as illustrated in  Fig. \ref{rot_2D_test_data_exa}.
Different from subsection \ref{subsec:regu2Dtest}, however,
random rotation and scaling within range from $0.5$ to $1.5$
is also applied when generating the model point sets
so as to test a method's ability to cope with arbitrary similarity transformations.

\begin{figure}[ht]
\centering
\includegraphics[width=1\linewidth]{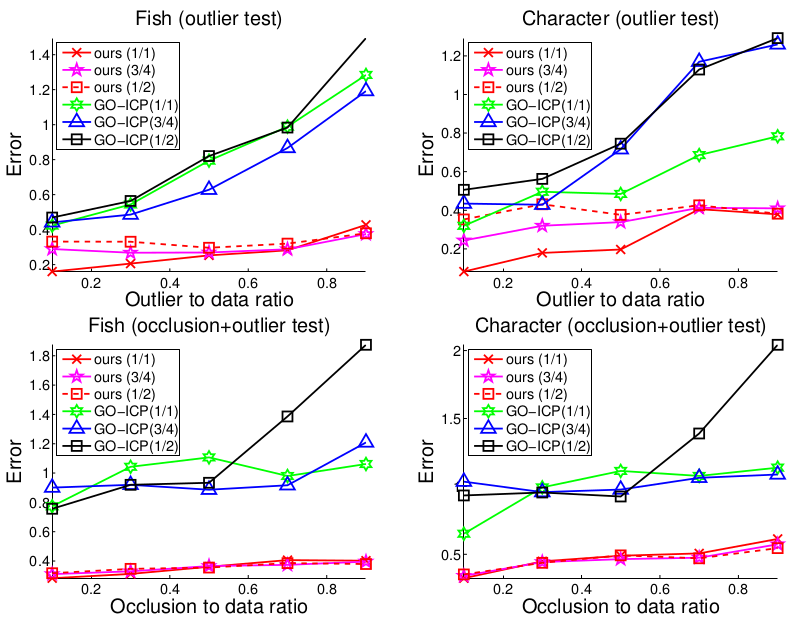}

\caption{Average  matching errors by our method 
and Go-ICP with different $n_p$ values (chosen from $1/2$ to $1/1$ the ground truth value)
over 100 random trials
for the 2D outlier and occlusion+outlier tests.
\label{2D_simi_sta}}
\end{figure}

The average matching errors  of our method and Go-ICP are shown in Fig. \ref{2D_simi_sta}.
It can be seen that our method performs much  better than Go-ICP,
especially for the occlusion+outlier test,
where there is a large margin between the errors of the two methods.
This demonstrates  robustness of our method to disturbances.

\begin{figure}[ht]

\includegraphics[width=1\linewidth]{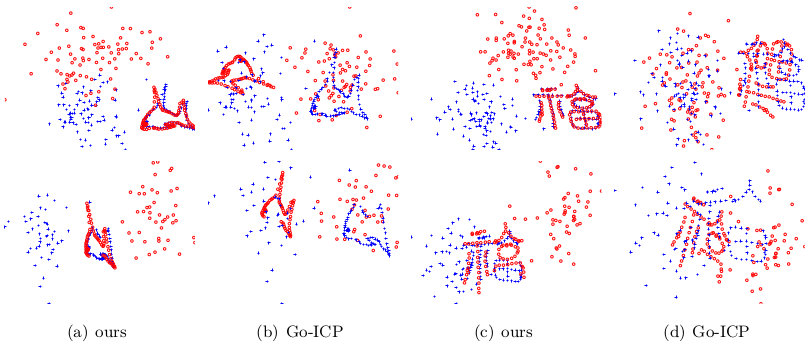}

%
%

  \caption{
Examples of   matching results by our method and Go-ICP 
for the 2D outlier (top row) and occlusion+outlier (bottom row) tests.
  \label{3D_nonrotate_exa}}
\end{figure}

The average running times of our method and Go-ICP are  1.3976 and 1.1701 seconds, respectively.

\subsubsection{ 3D synthesized datasets}

\begin{figure}[ht]
\centering

\includegraphics[width=1\linewidth]{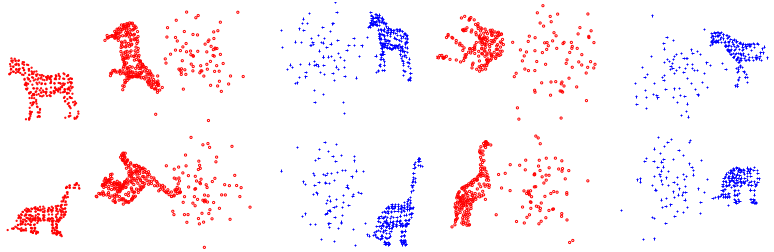}

\caption{
Left column: the prototype shapes.
For the rest columns:
examples of model and scene point sets in the 
outlier (columns 2, 3) and occlusion + outlier (columns 4, 5) tests. 
\label{rot_3D_test_data_exa}}
\end{figure}

Analogous  to the experimental setup in the previous   subsection,
we use two categories  of tests to evaluate  performances of different methods:
1) \textbf{Outlier test} and
2) \textbf{Occlusion + Outlier test},
as illustrated in  Fig. \ref{rot_3D_test_data_exa}.
The same two 3D prototype shapes as used in subsection \ref{subsec:regu3Dtest}
are used as the prototype shape, respectively.

\begin{figure}[!h]
\centering

\includegraphics[width=1\linewidth]{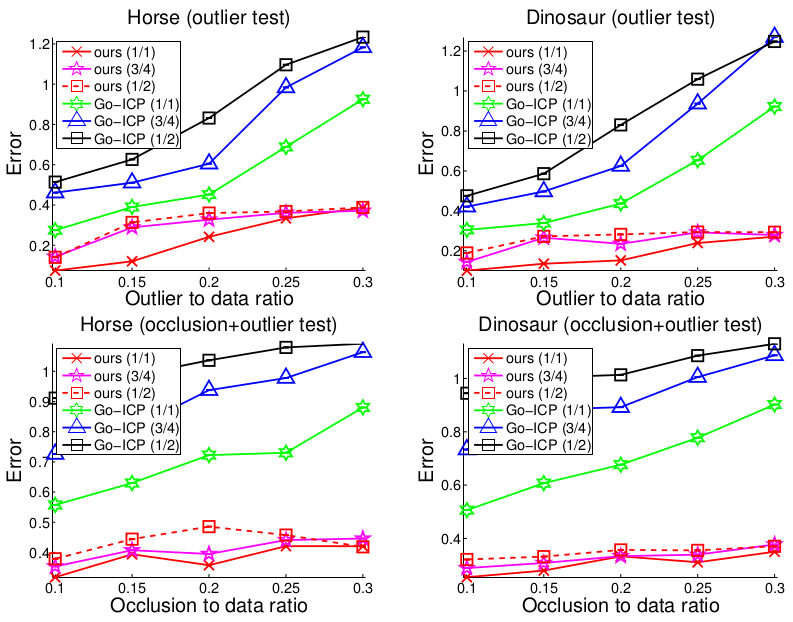}

\caption{
Average  matching errors by our method 
and Go-ICP with different $n_p$ values (chosen from $1/2$ to $1/1$ the ground truth value)
over 100 random trials
for the 3D outlier and occlusion+outlier tests.
\label{3D_simi_sta}}
\end{figure}

The average matching errors of our method and  Go-ICP are shown in Fig. \ref{3D_simi_sta}.
It can be seen that
our method performs much  better than Go-ICP,
especially for the occlusion+outlier test,
where there is a large margin between the errors of the two methods.
This demonstrates  robustness of our method to disturbances.


\begin{figure}[!h]

\includegraphics[width=1\linewidth]{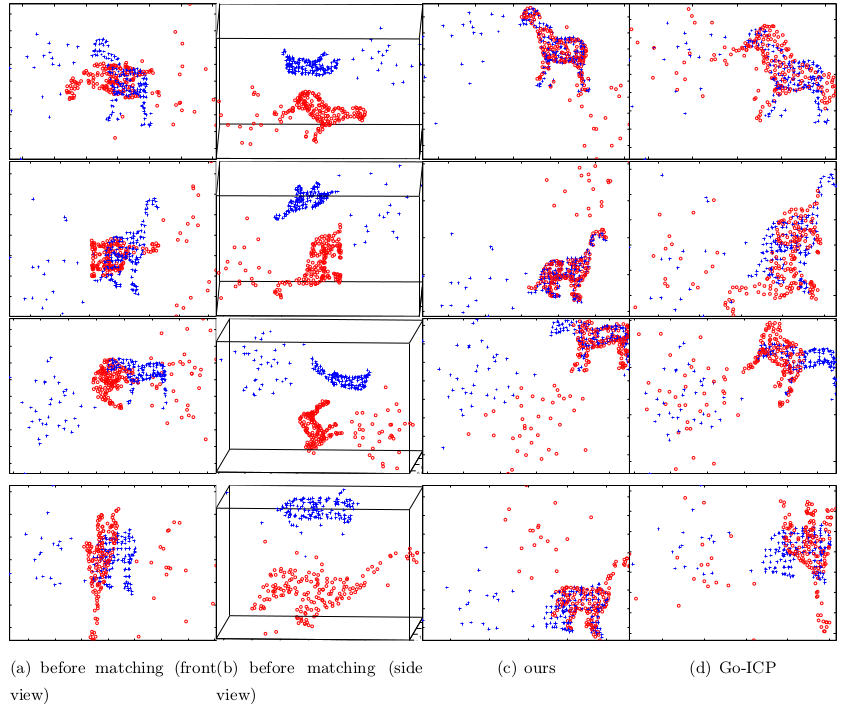}

  \caption{
Examples of   matching results by our method and Go-ICP 
in the 3D outlier (top row) and occlusion+outlier (bottom row) tests.
  \label{3D_nonrotate_exa}}
\end{figure}

The average running time of our method and Go-ICP are 152.7024 and 
    1.8105 seconds, respectively.

\section{Conclusion \label{sect:conclude}}
We proposed a new point matching algorithm capable of handling  the case that
there is only partial overlap between two point sets in this paper.
Our algorithm works by
reducing the objective function of RPM 
to a concave function of point correspondence
with  a low rank structure.
The BnB algorithm is then used for optimization.
Two cases of transformation,  
the transformation is linear with respect to its parameters and  the  2D/3D similarity transformations,
are discussed for our algorithm.
We also proposed a new lower bounding scheme
which has a k-cardinality linear assignment formulation
and can be very efficiently solved.
The resulting algorithm 
is approximately globally optimal,
scales well with problem size and is efficient for the 2D case.

Experimental results on both 2D and 3D datasets 
showed that the proposed method has strong robustness against disturbances
and outperforms state-of-the-art methods in terms of robustness to outliers and occlusions
with competitive time efficiency.


{\small
\bibliographystyle{ieee} 
\bibliography{../DP_SC_rotate/DP_SC_CVPR}
}

\end{document}